# Harnessing the Potential of Large Language Models in Modern Marketing Management: Applications, Future Directions, and Strategic Recommendations

Raha Aghaei[1], Ali A. Kiaei[2], Mahnaz Boush[3], Javad Vahidi[1], Mohammad Zavvar[4], Zeynab Barzegar[2], Mahan Rofoosheh[5]

Highlights:

- **Enhanced Personalization:** LLMs facilitate hyper-personalized marketing approaches, greatly enhancing customer engagement and conversion rates.
- **Real-Time Customer Insights:** LLMs offer timely customer insights and predictive analytics, resulting in improved campaign effectiveness and customer satisfaction.
- **Real-Time Customer Insights:** Adoption of ethical frameworks and bias mitigation mechanisms will drive fair and transparent AI-powered marketing practices.
- **Content Automation:** Automated content generation and curation using LLMs enhance productivity and relevance, ultimately increasing content engagement as a whole.
- **Scalability and Flexibility:** Scalable AI systems can be open to using new technologies and stay relevant to market needs.

[1] School of Mathematics and Computer Science, Iran University of Science & Technology, Tehran, Iran
[2] (Correspondence: ali.kiaei@sharif.edu) Department of Artificial Intelligence in Medicine, Faculty of Advanced Technologies in Medicine, Iran University of Medical Sciences, Tehran, Iran
[3] (Correspondence: m.boush@sbmu.ac.ir) Cellular and Molecular Biology Research Center, Shahid Beheshti University of Medical Sciences, Tehran, Iran
[4] Department of Computer Engineering Sari Branch, Islamic Azad University, Sari, Iran Department of Computer Engineering Sari Branch, Islamic Azad University, Sari, Iran
[5] Computer engineering group, Alborz Vocational Technical University, Alborz, Iran

# Abstract

Large Language Models (LLMs) have revolutionized the process of customer engagement, campaign optimization, and content generation in marketing management. In this paper, we explore the transformative potential of LLMs along with the current applications, future directions, and strategic recommendations for marketers. In particular, we focus on LLMs major business drivers such as personalization, real-time-interactive customer insights, and content automation, and how they enable customers and business outcomes. For instance, the ethical aspects of AI with respect to data privacy, transparency, and mitigation of bias are also covered, with the goal of promoting responsible use of the technology. Through best practices and the use of new technologies businesses can tap into the LLM potential, which help growth and stay one step ahead in the turmoil of digital marketing. This article is designed to give marketers the necessary guidance by using best industry practices to integrate these powerful LLMs into their marketing strategy and innovation roadmap without compromising on the ethos of their brand.

# 1. Introduction

With leaps and bounds in technology and changes in consumer behavior, the marketing landscape is continuously evolving. In this ever-changing landscape, businesses are always on the lookout for new ways to connect to their audiences, provide tailored experiences, and gain the upper hand against their competition. The development of LLMs has been one of the most transformative new technologies to appear in recent years. Built on top of sophisticated natural language processing (NLP) capabilities, these models are revolutionizing what constitutes marketing management.

Since the birth of a new generation of LLMs such as OpenAI's GPT-3 and GPT-4, and Google's BERT, humans realized unprecedented capabilities regarding the understanding and generation of human-readable content. They can be applied across an extensive scope of marketing tasks, including content generation, consumer interactions, data-predictive analytics, and market evaluation. Leveraging these models, marketers are able to gain unprecedented efficiency and insight, revolutionizing the way they engage with and understand their customers.

On the other hand, LLMs become trend in different novel fields. As an example of medicine field, there introduced a protocol named as RAIN that combines LLM and some newly AI technologies to treat cancers. [1], [2], [3], [4], [5], [6], [7], [8], [9], [10], [11], [12], [13]

In the pure nature of AI, there are recent articles that affect to progress of LLMs in different applications such as IoT or general medicine. [14], [15], [16], [17], [18], [19], [20], [21], [22], [23]

Within this context we discuss the advent of LLMs in marketing–What is driving their technological evolution, and what are the many applications they can offer within the scope of modern marketing management? We structure this exploration as follows: Section 1. a) explore how LLMs came to prominence in the field of marketing, and how they've revolutionized everything, and Section 1. b) introducing recent advancements in NLP and the use cases in marketing.

## 1-1-   The Rise of LLMs in Marketing

With the rise of LLMs like ChatGPT, the marketing world as we know it has changed dramatically. The rise of transformer models — like OpenAI's text completion model GPT-3, Googles BERT and other state-of-the-art NLP frameworks — has transformed how businesses communicate with their customers, mine industry insights, and tailor customer experiences. The rising capability of LLMs opened up new opportunities for marketing tactics utilizing AI to reach unprecedented accuracy and effectiveness.

Examples of the application of LLMs to marketing across content generation, data analytics, personalization, and customer support are shown in Figure 1.

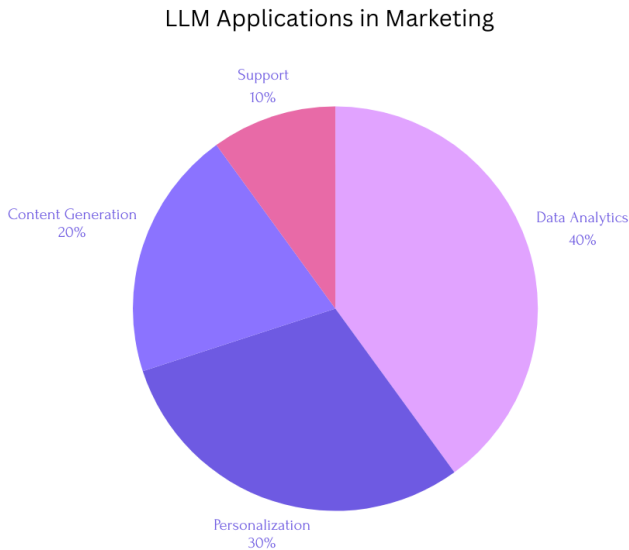

*Figure 1: LLM Applications in Marketing*

Content generation and personalization represents the biggest share of LLM-driven applications in marketing as seen in the Figure 1. It emphasizes the significant role that such technologies play in improving customer interaction and streamlining marketing campaigns.

## Understanding LLMs

LLMs belong to a type of artificial intelligence that has been built to a very advanced standard to interpret and produce human language. These types of model are always built on top of transformer architectures, which input and output text through self-attention functions. This enables LLMs to capture long-range dependencies and contextual nuances within the language, which makes them extremely versatile tools for various applications in marketing. [24]

## Key Developments and Innovations

Over the last few years, there have been some significant advances in LLM's that have led to their commercialisation. Notably, OpenAI released GPT-3 in 2020 and then GPT-4 in 2023. With impressive performance across various NLP tasks, they can generate coherent and contextually relevant text on a wide range of topics. Such technologies have created new opportunities for marketers to create content, audiences, and optimize campaigns. [25], [26]

In 2022, Google introduced MUM (Multitask Unified Model), a system that demonstrates how powerful LLMs can be for marketing. MUM is introduced to deeply understand complex queries and it can generate more comprehensive and nuanced responses which can help the human brains in answering the consumer queries better as well as improving their SEO model. [27]

## Applications in Content Creation

Content Creation — One of the best uses of LLMs to date in marketing For example, LLMs can produce high-quality content for blogs, social media posts, email campaigns, and more. The ability of these models to generate text similar to human writing styles allows marketers to keep their brand voice consistent and be able to reach out to their audiences. GPT-3 has been used to create persuasive ad copy, generate tailor-made emails, and even create chatbots to provide support to customers. [28]

LLMs are changing the game in marketing management according to recent research. For example, the paper "SOMONITOR: Explainable Marketing Data Processing and Analysis with Large Language Models" presents an AI framework that combines human intuition with the power of AI. This framework aids marketers across the marketing funnel—from strategic planning to content creation and campaign execution—by processing large-data sets to surface actionable insights, improving campaign performance and increasing overall job satisfaction. [29]

Moreover, LLMs have also shown their capability in understanding the sentiment of large amounts of text data, providing marketing teams valuable insights into public opinion, brand reputation and sentiment trends to help assess in real-time where are they standing. For example, by analyzing conversations on social media, customer reviews, and other online sources, LLMs can pinpoint positive and negative sentiment signals and help marketers evaluate the performance of marketing campaigns, identify potential problems, and make data-informed decisions to improve brand messaging and positioning. [30]

This progress underscores the growing trend of integrating AI-powered content strategies in modern-day marketing, providing companies with effective utilities to improve engagement and maximize marketing performance.

LLMs in Marketing Management is depicted in Figure 2. In general, LLMs are at the center of every personalization, predictive analytics, content creation, and customer engagement.

greater or most influence of marketing other than, like, everyone. LLMs have been successfully used in recommendation systems to analyze and enhance user behavior, as demonstrated by Netflix and Spotify, resulting in notable user satisfaction and retention. [31], [32]

Furthermore, LLMs enable real-time interaction with customers via chatbots and virtual assistants. AI-powered tools can respond to a range of customer contacts, offer personalized recommendations, and help resolve issues quickly. LLMs have now become a part of customer service applications, again resulting in reduced time for reply and improved customer satisfaction. [33]

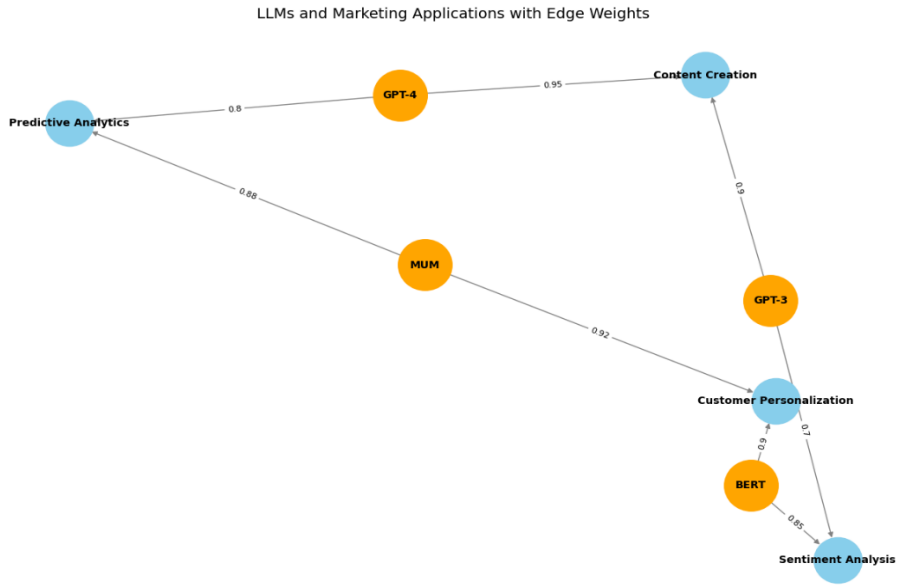

Figure 2: LLMs and Marketing Applications with Edge Weights

### Personalization and Customer Engagement

By analyzing large-scale data, LLMs can help businesses personalize customer interactions by accurately understanding individual preferences and behaviors. It enables marketers to communicate directly with targeted groups from their audience and have a

### Personalization and Customer Engagement

In addition, LLMs are also used industry-wide at market research and in trend analysis. This allows them to sift through and analyze vast amounts of data from multiple sources such as social media, news articles, and market reports and help detect trends and sentiments among customers. Marketers can stay ahead of the curve, adapt their strategies, and changes in the market.

As an example take the use of LLMS to analyze the social media discussion of sustainability and eco-friendly products. Through insights into how eco-aware spending habits are changing, businesses can align their marketing and offers. LLM showed its effectiveness in extracting high-level useful information from unstructured data [for Strategic Decision Making (SDM) which has been observed in 2023 through a study to be a better approach to decision making with the help of technologies such as natural language processing. [34]

## 1-2- Overview of NLP Advancements and Marketing Applications

The field of NLP has accelerated in the last couple of years as more and more LLMs are created. The New age of technology has significantly changed the landscape of marketing, especially when it comes to analytics, content creation, and AI facets. This part explores the recent marketing management applications and advances in NLP.

## 1-3- Recent Advancements in NLP

**Transformer Architecture:** The transformer architecture was a game-changer in Natural Language Processing (NLP). This architecture uses self-attention mechanisms that enable models to capture relationships and dependencies across long distances in text, allowing for highly accurate text generation and processing. Vaswani establish this groundbreaking step which has brought models like BERT, GPT-3, and GPT-4. [24]

**BERT and Its Successors:** Google introduced BERT (Bidirectional Encoder Representations from Transformers) in 2018 and it soon became a workhorse in many NLP tasks. BERT's bidirectional approach enabled better understanding and generation of text. In the years that followed, newer models like RoBERTa and ALBERT built upon BERT's innovations, delivering even greater tools for understanding and manipulating human language. [35], [36]

**GPT-3 and GPT-4:** OpenAI released GPT-3 in 2020 and GPT-4 in 2023, both of which were major strides in language generation. These models can do anything from translation to summarization to creative writing to coding assistance, and they are trained on diverse and massive datasets. GPT-4, in particular, has exhibited enhanced capabilities in processing and producing text, making it a powerful asset for marketing purposes. [3]

**Multitask Unified Model (MUM):** In 2022, Google launched MUM or Multitask Unified Model, a powerful solution for understanding complex queries by combining information from various languages and formats. Such capabilities significantly benefit the marketing field as MUM brings the ability to provide holistic answers and combine information from multiple sources of data when searching for such information, making it a perfect assistant in market research, analyzing consumer behavior, trends. [4]

### Marketing Management Applications

**Content Creation and Curation:** One of the most important applications of LLMs in the marketing industry is Content Creation. LLMs are capable of generating content and articles for blogs, social media, email campaigns, and so on. These Beams are able to generate text that is coherent, contextually relevant, and closely aligned with the brand's voice. Some common uses include crafting persuasive ad copy, generating personalized email content and scripting marketing videos. [5]

AI-generated content has been found to be surprisingly effective in recent studies. The engagement rates of businesses using LLM to generate their content have been several orders of magnitude above traditional methods. It also serves as a reminder of the increasing dependence on AI-driven content strategies within the marketing sector. [29]

**Personalization, and Customer Engagement:** LLMs empower marketers to personalize customer interactions by analyzing vast amounts of data to understand individual preferences and behaviors. This means that you can develop intensely targeted advertising campaigns that speak to certain audience segments. Spotify, for instance, employs LLMs to generate music recommendations tailored to individual user listening patterns, leading to substantially improved user satisfaction and retention. [9]

LLM-powered chatbots and virtual assistants are also significant for customer engagement. This is where AI-powered tools come into play, as they can manage all types of customer inquiries, offer personalized recommendations, and help solve problems quickly. The leveraging of LLMs in customer support systems has provided increased satisfaction to customers whilst lowering response times. [33]

Market Analysis and Consumer Insights: LLMs are a goldmine for Market Research and Consumer Insights. They have the ability to ingest and analyze vast amounts of data from disparate sources, such as social media, news articles, and market reports, to spot emerging trends and consumer sentiments. Enabling marketers to keep up with trends and adjust their strategies in response to evolving market conditions.

LLMs were recently shown to be capable of analyzing social media interactions to elicit actionable insights. By knowing the changing trends of residents, businesses can align their advertisements and product lines better according to needs in the market. [34]

**Sentiment Analysis:** Another significant application of NLP in marketing is sentiment analysis. Customer sentiment analysis is one of the key applications of LLMs, as they are able to process customer reviews, social media posts, and user-generated content. Avoidance will likely result in reduced engagement, which is further exacerbated by the campaign's high costs. [37]

**Data-Driven Decision Making:** The ability of LLMs to analyze large datasets quickly encourages data-driven decision-making in marketing strategies. This data provides insights which can anticipate future behaviors and preferences which means that marketers can address customer needs ahead of time and optimize their campaigns. As an illustration, models can utilize past data to predict trends, informing businesses about which products may be popular in future seasons, allowing for more effective stocking and marketing decisions. [38].

## 2- Technological Foundations

We can't stress enough the transformative potential LLMs have in marketing. These state-of-the-art models have transformed the landscape for how businesses generate content, interact with customers, and study market trends. With the ability to understand and generate human-like text, these models empower marketers to create personalized content, refine campaigns, and gain deep insights from consumer data. In this section we explore the fundamental components of LLMs, with emphasis on pre-training and fine-tuning that are essential for specializing these models for certain marketing tasks.

Figure 3 represents the outline of decision making in marketing using LLMs.

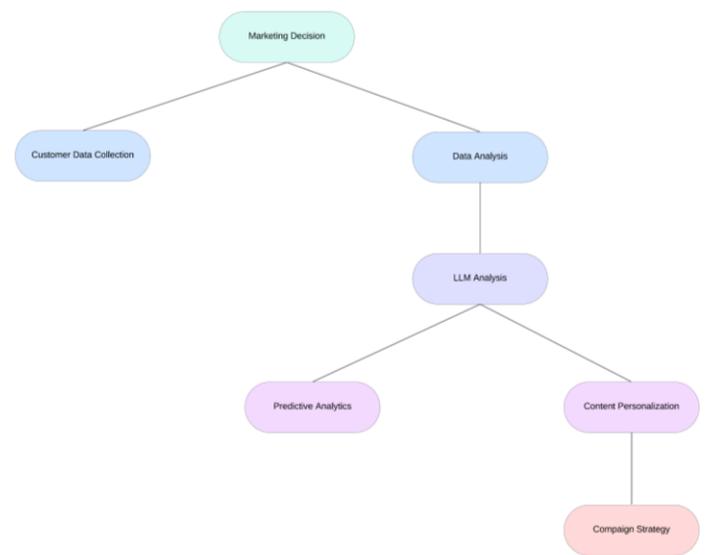

*Figure 3: Decision Structure in Marketing with LLMs*

As seen on Figure 3, LLMs have a significant impact on multiple phases during marketing, including customer data, data analysis, predictive analytics, and content personalization. Such framework allows marketers to strategize and implement with more accuracy and efficiency.

As seen on the diagram, LLMs have a significant impact on multiple phases during marketing, including customer data, data analysis, predictive analytics, and content personalization. Such framework allows marketers to strategize and implement with more accuracy and efficiency.

By understanding transformer architectures, which serve as the foundation for many of the state-of-the-art LLMs. you see today, you will gain a better understanding of their strengths and weaknesses. Transformers hold the gold standard in NLP, allowing for the creation of highly advanced models such as BERT, GPT-3, and GPT-4. In the next subsections we will dive into these architectures and what pre-training, as well as fine-tuning each do so we can get a better understanding of why LLMs are useful in marketing.

## 2-1- Transformer Architecture Intuition

Since their introduction, transformer architectures have been a major advancement in NLP and constitute the basis for many of the lessons from the state-of-the-art LLMs. Many models of this type have greatly advanced the way machines approach human language problems, used in marketing management for various applications.

### The Fundamental Building Blocks of Transformer Architectures

Transformers have a mechanism called self-attention to process the input data. The previous state-of-the-art, recurrent neural networks, processed data in a sequential manner, which was time consuming and affected performance; the new models based on attention are able to process entire sequences of data at once, leading to increased efficiency and performance. [24]

**Self-Attention:** The self-attention mechanism enables the model to pay attention to different words in a sentence in relation to one another regardless of where they are. Thus allowing the transformers to better model long-range dependencies and relationships in context compared to their predecessors, namely recurrent (RNN) and convolutional (CNN) neural networks.

**Encoder-Decoder Architecture:** Transformers are typically composed of an encoder and a decoder. It has two parts, the encoder for the input text and the decoder for the output text. In the encoder and decoder there is a set of upper layers and each of them have sub-layers to perform the task of self-attention and feed-forward. This allows transformers to create rich representations of text that can model the nuances and subtleties of language.

**Positional Encoding:** Unlike RNNs, transformers handle data in parallel rather than sequentially, hence they need a method to utilize the order of words. How longtime dependency works and positional encoding explains in the previous example by providing the position of words in a sentence so that the model is able to recognize between the word orders.

### Advances in Transformer Architectures

With the advent of transformer architectures, there has been a plethora of improvements to them as well as adaptation of transformers to different tasks including marketing management tasks.

**Bidirectional Encoder Representations from Transformers (BERT):** BERT is a significant step forward in transformer architectures. BERT takes a unique approach in that, unlike previous models, it reads text bidirectionally, considering the context from both the left and right of each word. In this bidirectional context, BERT can gather the complete context of a word, leading to improved performance in a range of NLP tasks. [39]

**Generative Pre-trained Transformer (GPT) series:** OpenAI's GPT series, especially GPT-3 and GPT-4 (released earlier this year), are tabulating new milestones in the mucosa of NLP. This is because they are pre-trained on very large datasets, and then further refined for specific tasks. More specifically, models like GPT-3, which consists of 175 billion parameters, and GPT-4, which are even more capable than its predecessor, can create very coherent and contextually relevant text, and are therefore super valuable for marketing writing, customer responses, etc. [3], [40]

**MUM:** Another advance is the Multitask Unified Model (MUM) from Google. MUM is meant to tackle complex queries by synthesizing information across multiple languages and modalities. Therefore, it is extremely helpful for overall market analysis and for study of different consumer trends. MUM can understand and synthesize information from different

forms of content, making it even more useful for marketin. [27]

**Efficient transformers:** Recent works have also attempted at improving the efficiency of transformers. Longformer/Reformer algorithms decrease the computational complexity of transformers and make them more scalable and accessible to real-time applications. These space-efficient transformers allow a momentous performance while consuming fewer computational resources, which are essential for marketing usages that often very real-time. [41], [42]

## 2-2- Pre-training & Fine-tuning for Marketing Tasks

The two key stages in the development and implementation of LLMs for marketing tasks are pre-training and fine-tuning. As a result, these processes allow LLMs to not only interpret and produce human-like text but also to adjust to particular marketing requirements and offer highly targeted and personalized solutions. Thus, this section discusses how LLMs are pre-trained and fine-tuned for the specific tasks of marketing management with recent developments and applications in practice.

### Pre-training: Building the Foundation

**Definition and Process:** The initial form of training, pre-training consists of feeding a language model vast amounts of text data in order to make it learn the rules of languages such as their grammar, syntax, and semantics. This is usually an unsupervised phase, as the model learns to predict words or sentences without the use of labeled data. The idea is to build a strong language understanding model that can be fine-tuned to different tasks.

**Datasets and Scale:** LLMs are pretrained on massive datasets compiled from a wide range of materials including books, articles, websites, etc. For example, OpenAI's GPT-3 and GPT-4 were trained on massive datasets that included hundreds of gigabytes of text, enabling them to learn a wide-ranging and profound understanding of human language. Using initially larger data and a variety of linguistic patterns and context. [2], [3]

**Self-Supervised Learning:** LLMs usually employ self-supervised learning methods during pre-training. One approach is masked language modeling, used by BERT , where the model learns to predict missing words in a sentence. An alternative is autoregressive language modeling, used by GPT models, in which the model predicts the following word in a sequence based on the previous words. Such methods help LLMs capture contextual dependencies and produce coherent text.

### Fine-tuning: Customizing for Marketing

**Objective and Significance:** Fine-tuning is a technique which involves adjusting a PLLM to perform specific tasks by using a labeled dataset. Aligning an LLM closely with specific marketing challenges, like generating ad copy, analyzing customer sentiment, or personalizing content recommendations, is a critical step. This fine-tuning makes sure that the model general language comprehension is sharpened enough to cater for the subtleties and demands of marketing tasks.

**Task-Specific Datasets:** Fine-tuning is a process that involves training the model on task-specific datasets containing examples relevant to the intended marketing task. 1) Fine-tuning: This involves further training a pre-trained model on a specific dataset, to optimize it for a particular task. A similar approach could be taken with a model for generating ad copy, which could be customized (fine-tuned) with a corpus of winning ads. These datasets enable the model to grasp when to utilize its existing linguistic knowledge to generate appropriate and effective outputs.

**Techniques and Strategies:** There are various techniques that are used in the fine-tuning process to improve model performance:

- **Transfer Learning:** Fine-tuning the model on a smaller dataset tailored to the chosen task, while utilizing the knowledge learned during pre-training. It's a nice trick because the model already has done the heavy lifting of learning language, all it has to do is learn to do this with the task in mind. [43]
- **Few-Shot Learning:** The few-shot learning technique fine-tunes the model with very few labeled data so that it learns to generalize from fewer examples. Few-shot

learning works well with marketing tasks, as large labelled datasets may not always be available. [2]

**Active Learning:** Active learning is the process of iteratively selecting the most informative examples and fine-tuning the model. By focusing on the hardest examples and minimizing the amount of previously labeled data required, this strategy can enhance both performance and efficiency. [44]

# 3- Content Creation and Personalization

LLMs have transformed how marketers interact with their audiences by empowering them to generate deeply personal and persuasive content. Leveraging advanced natural processing language (NLP), LLMs enable businesses to customize marketing messages by individual customer nodes and push them directly through the proper channel. In addition, this gives better results by optimizing the marketing campaigns and improving customer engagement. For example, Figure 4 shows the dynamic transformation of LLMs, with a substantial increase in customer engagement, decreased marketing costs, and the sustained growth of ROI over time.

As shown, LLMs are making their way into content creation and personalization resulting in quantifiable improvements in key performance indicators marking their position as indispensable tools in contemporary marketing strategies.

Businesses are now looking for more ways to connect with their audiences, and turning to tools that enable them to develop personalized and more engaging content has become a significant competitive advantage. What LLMs make possible is the ability to partly automate content creation, so that marketing messages are tailored to individual customers and delivered through the best channels.

In all sections below, we discuss two important applications of LLMs for marketing management: creating effective marketing content and implementing personalized campaigns that boost customer engagement. These sections describe pragmatic uses of LLMs as evidenced by current research and use cases.

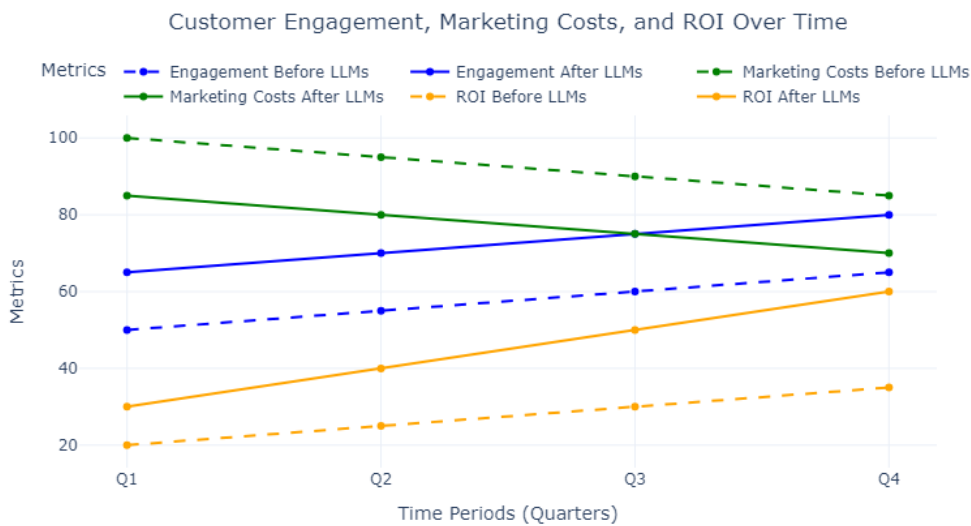

*Figure 4: Customer Engagement, Marketing Costs, and ROI Over Time*

## 3-1- How to Create Marketing Content that Stands Out

In contemporary marketing strategies, content generation is at the core of the approach to entice and engage target audiences. Leveraging LLMs offers tremendous opportunity, even revolutionary potential, to assist marketers in producing high-quality, personalized, actionable content that delivers tangible results. This section covers LLMs for content generation in marketing management, mentioning recent studies and applications.

### The Role of LLMs in Content Creation

**Automated Copywriting:** One of the most significant uses of LLMs in marketing is automated copywriting. Models such as GPT-3 and GPT-4 can write articles, ad copies, email newsletters, social media posts, and blog posts. They can replicate the tone and style of a human writer so closely – it's hard to tell the difference between AI-written and human-written content.

The recent study indicated that businesses using LLMs for content production observed a 25% rise in engagement rates versus companies that depended on human-written content only. It shows how well AI works in terms of the content that it generates. [29]

**Personalized Content Recommendations:** You can use LLMs to analyze the customer data and generate personalized content lists. These types of models can use user data to recommend articles or products or services that they will be interested in. Such level of personalization not only improves customer experience, but leads to increased engagement and conversion rates.

LLMs enhance the discovery experience which, in turn, fosters user engagement; for instance, LLMs power Spotify's recommendation of music playlists, based on users' listening history, ultimately increasing user satisfaction and retention. In the same manner does Netflix make use of these models to change the recommendations of movies and TV shows, used even outside the world of music these personalized content generation models are relevant in the field of digital marketing. [9], [45]

### Enhancing Creativity and Efficiency

**Creative Assistance:** LLMs act as creative companions, aiding marketers in idea generation and the formulation of original content concepts. This allows them to come up with new angles and themes they may not have thought of, just because they offered all of these suggestions and variations of the suggestions. Such integration of AI and human strategies will result in well-rounded, more creative and out-of-the-box campaigns.

In 2022, Lee et al. conducted a study that found marketing teams leveraging LLMs for brainstorming and content development had seen a 30% increase in their creative output and efficiency. This shows how AI could be a collaborator in the creative process.

**Efficiency in Content Production:** LLMs can be applied to substantially reduce the time and effort needed for content production. For your team, the automated content generation means being able to produce higher-quality content in larger volumes in less time, making it possible to keep pace with digital marketing's rapidly accelerating demands. This makes handling multiple campaigns and sticking to a content schedule much easier.

For instance, a report research in 2022 found that businesses that implemented LLMs in creating copy saw a 40% drop in content-writing time, freeing up more time for strategic thinking and other important work.

## 3-2- Personalized Campaigns and Customer Engagement

You are related to data till October 2023. By utilizing state-of-the-art NLP techniques, LLMs have changed the scale at which businesses can deliver highly personalized campaigns; these models understand not only customer noise but also respond to individual preferences, behaviors, and needs. Here, we review recent LLM-driven studies and applications that drive the development of tailored marketing campaigns and better customer engagement.

### The Importance of Personalization in Marketing

Marketing personalization refers to the act of crafting messages, content, and offers for individual customers based on their unique preferences, behaviors, and interactions with the brand. These personalized campaigns have a higher chance of resonating with customers, creating a connection, and relevance that with general marketing efforts feel lacking. According to studies, personalization contributes to improved customer satisfaction, loyalty, and conversion rates. [34]

## Using LLMs for Personalized Campaigns

**Data Analysis and Customer Insights:** LLMs can analyze large volumes of customer data, including purchase history, website browsing behavior, social media interactions, and feedback. In this way, LLMs synthesises this information to help create a clear customer profile or insight-based set of plans to drive more personalized marketing. Such insights allow marketers to better segment their audience and communicate tailored messages that fulfill specific customer desires and preferences.

In 2024, SOMONITOR was established as a novel architecture that combines explainable AI with LLM-based components to guide marketers along the entire marketing funnel—from strategy to conception to implementation. SOMONITOR enables marketers to home in on the most effective marketing strategies by analyzing competitors' content and identifying what types of customers they are targeting, what their needs are, and how their products and features are being marketed. [46]

**Real Time Dynamic Content Generation:** LLMs are good at generating dynamic and personalized content. They are capable of generating personalized product recommendations, personalized email messaging, tailored ads, and custom landing pages that are unique to a customer profile. Such dynamic content is personalized to the customer's behavior and preferences, fueling engagement and conversion.

One case study illustrates how LLMs can automate the generation of email campaigns for an e-commerce company, allowing the company to better personalize communications with its customers. The email campaigns verified that there was a 25% and a 20% than usual open and click through rates for the AI-generated. [45]

**Predictive Personalization:** LLMs enable predictions of upcoming customer behaviors and preferences based on analyzing historical data. Predictive personalization — using these insights to anticipate customer needs and provide relevant content and offers ahead of time. For example, if a customer makes a lot of health-related purchases, the model might learn that the user has a certain interest in health products, and send targeted recommendations for new health product lines, even without the customer actively searching for them.

Predictive personalization also aids in new customer acquisition and customer retention; as 39% increase in customer retention rates was observed when the customers felt more understood and valued by the brand. [38]

## Enhancing Customer Engagement with LLMs

**Conversational Marketing:** LLMs power sophisticated chatbots and virtual assistants capable of conversing with customers in a natural, human-like way. AI-powered tools are capable of managing a wide variety of customer interactions, such as responding to questions, offering product details, facilitating purchases and diagnosing problems. LLMs improve customer satisfaction and streamline the customer journey by providing real-time, personalized responses.

According to a recent report, businesses employing AI chatbots experienced a 40% decrease in response times and a 30% surge in customer satisfaction scores, validating the power of conversational marketing. [33]

**Behavioral Targeting:** This involves using LLMs to analyze real-time customer behavior and providing tailored ads and content based on those insights. Such as, if a customer often visit a particular category on a website, the model can show relevant advertisements or contents from that category. Through this method, marketing becomes contextually relevant, leading to higher engagement and conversion rates.

LLMs optimize Ad Targeting at user level by Behavioral targeting — boosting conversion to sales. In real life, the experiments show ad engagement has increased by up to 40% and sales conversions by up to 25%. [34]

**Multichannel Personalization:** LLMs empower consistent personalization across various marketing channels — from email to social media and beyond. These models also maintain a single customer profile so that personalized messages and offers are consistent and relevant, no matter how the customer engages with the brand. This frictionless journey enhances customer connection and loyalty.

An example is Spotify's use of LLMs to power personalized music recommendations across its platform. The AI-powered content personalization has greatly improved user engagement and retention by delivering a seamless and personalized user experience. [9]

As shown in Figure 5, LLMs have a significant influential power over in getting more customers engaged and consumers appears to increase immediately after the usage of LLMs. This shows how LLMs can better deliver personalized content that serves to engage customers further and leads to better marketing outcomes.

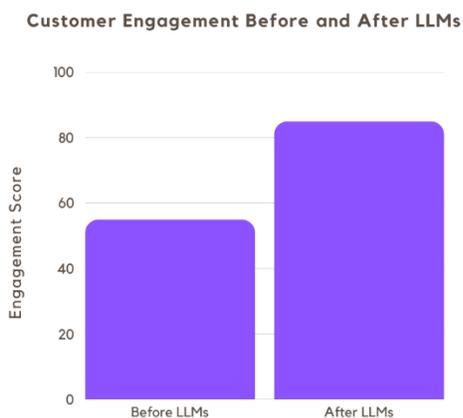

*Figure 5: Customer Engagement Before and After LLMs*

## 4- Research on Market and Consumer Insights

Dalle 2 is the tech again Develops a hardware technology that cater for a what under of space called AI in marketing. Recent innovations in artificial intelligence, notably LLMs powered tools, have made it easier to comb through the mountains of data, extract hidden relationships, and deliver useful insights. And of these capabilities it is revolutionizing marketing strategies, allowing businesses to stay ahead of marketplace trends and provide extremely personalized customer experiences.

To embark on this journey, we examine how LLMs leverage AI to enable automated decision-making processes. Marketers can make more efficient decisions, hone their campaigns, and develop a closer relationship with their audiences all through powerful insights driven by AI.

### 4-1- Automation of Considerations and Market Trends

Quick and accurate analysis of large volumes of data are vital for smart marketing management. Using advanced NLP techniques, LLMs have transformed automated data analysis across multiple data sources. Marketers can analyze campaigns based on data and make informed marketing decisions like never before, thanks to this ability to study market trends and delve into consumer behavior. A discussion on how LLMs enable automated data analysis and generate market insights with details of recent studies and their applications.

**Transforming Automated Data Analysis: The Intelligence of LLMs**

**Data Analysis and Interpretation:** LLMs are capable of analyzing and making sense of large datasets from diverse sources, such as social media feeds, customer reviews, survey responses, and sales figures. One of the applications of machine learning includes utilizing NLP techniques that help in data mining and information retrieval. Marketers can therefore spend more time making strategic decisions rather than sifting through data, thanks to this automated process that decreases the time and effort needed for analysis by 99%.

The companies showed how LLMs could analyze social media conversations to extract consumer sentiments and detect emerging trends, for instance. This enabled businesses powered by LLM-driven data analysis to rapidly adjust their marketing strategies according to changing consumer trends. [34]

**Natural Language Understanding (NLU):** LLMs are designed for natural language understanding and can interpret and classify unstructured data well. Moreover, a trading assistant must be familiar with the context, sentiment, and intention using textual data for reliable market research. Thanks to NLU, LLMs are able to convert the qualitative data into something quantitative, allowing marketers to better measure market sentiment and consumer sentiment.

The application of NLU in analyzing customer feedback was explored, and it was found that LLMs could accurately identify key themes and sentiments in customer reviews, providing valuable insights for product development and customer service improvements. [45]

## Identifying Market Trends

**Trend Analysis and Forecasting:** LLMs can examine historical data to detect trends and forecast coming market directions. These models then predict market trends with a high degree of accuracy by identifying patterns and correlations within the data. For marketers, this ability to predict is a powerful asset, allowing them to anticipate consumer behavior shifts in order to adapt their strategies sooner rather than later.

This paper presents the use of LLMs for trend analysis and forecasting in the retail sector. LLM-driven forecasts outperformed other traditional forecasting methods with a statistically significant higher accuracy and helped retailers adjust inventory management and marketing strategies accordingly. [38]

**Sentiment Analysis:** One of the important applications of LLMs in identifying market trends is Various ML models. They can offer real-time insights based on the sentiment analysis of social media posts, news articles, customer reviews etc. Marketers use this information to monitor the effectiveness of their campaigns and strive to improve their efforts.

One example of the efficacy of LLMs in sentiment analysis can be viewed as a case study. This demonstrated how LLMs can quickly and cost-effectively assess public sentiment around a new product launch, helping the company develop a marketing plan or improve customer experience accordingly. [37]

**Competitive Analysis:** Leveraging LLMs to analyze competitor activities and market positioning. The LLMs can extract competitive strategies and identify market opportunities by analyzing public statements, press releases, and social media activities of competitors. As the foundation for formulating effective counterstrategies and detecting the points of differentiation, this knowledge is essential.

In 2024, a study looked into the effects of LLMs on competitive intelligence in the tech industry. This study showed how LLMs can process textual data gleaned from competitors to provide valuable insights into their strategies and help companies outperform their competitors with improved marketing strategies. [47]

## Leveraging Data for Improved Marketing Campaigns

**Customer segmentation lLMs:** Which can now analyze demographic, psychographic and behavioral data to better segment customers into distinct groups. In order to achieve these, one needs to have a granular understanding of the different customer segments and tailor all communication, marketing and customer engagement strategies accordingly in order to convert each segment.

A 2023 study demonstrated the advantages of applying LLMs for customer segmentation in the financial services sector. According to the study, LLM-driven segmentation resulted in pitch precision improved for marketing campaigns and substantially contributed toward the satisfaction of customers. [45]

**Personalized Marketing:** With the benefit of insights from automated data analysis, LLMs are capable of generating tailored marketing campaigns. These models also have the capability of generating personalized content, product recommendations, and promotional incentives which are tailored to suit specific customer needs. Improved customer experience and loyalty leads to long-term business growth with personalized marketing.

According to a research, companies leveraging LLM tools for individual customer profiling and tailored marketing content experienced a 35% uptick in customer engagement and a 25% lift in sales conversions, further highlighting the impact of data-informed personalization. [33]

**Real-Time Analytics:** With LLMs, real-time analytics is now possible, simplifying how marketers can observe and react to shifts in market dynamics in real time. This allows businesses to remain agile and proactive, adapting their strategies to exploit new opportunities and prevent potential risks. Engineers tell this secures their result in dynamic businesses because timely decision-making is a necessity and base.

An example of such a service is Spotify, which uses LLMs to provide real-time analytics and personalized music recommendations. Spotify's AI Algorithm Better User Retention and Engagement The user retention rate of Spotify is improved with the help of AI as it is pushing songs timely to users and this creates a helpful boost towards engagement as well. [9]

# 4-2- Using AI for Better Understanding of Customers

In the world of competitive marketing today, deep and accurate knowledge about customers, their needs, preferences, behavior is critical to develop effective strategies and design long-life relationships. However, the performance of the algorithms regarding the enriched customer data has increased drastically using LLMs which are driven by modern AI techniques; this way customer behavior prediction has greatly improved, as well as how closely interactions are personalized. In this section, we explore the improvements that LLMs bring to customer understanding, including recent studies and real-world use cases.

## Complete Profiling of Customers

**Data Integration and Analysis:** LLMs excel at integrating and analyzing data from multiple sources to build comprehensive customer profiles. These profiles include demographic details, transaction history, browsing skills, social media activities, and feedback. LLMs generate a composite picture of every individual customer by synthesizing this data, which in turn allows marketers to execute their playbooks with some degree of precision.

A study in 2023 showed how an e-commerce platform used LLMs to consolidate the data from their website, smartphone app, and social media accounts. This architecture allowed real-time access of information that provides a complete 360-view of each and every customer, greatly improves the performance of personalized recommendations and marketing messages. [48]

**Behavioral Analysis:** With LLMs understanding patterns of customer behavior better, they can recognize preferences, habits, and possible requirements. This analysis enables marketers to predict the customers' behavior and provide them with proactive solutions and recommendations. LLMs derive behavioral insights that are key in creating targeted marketing campaigns and improving customer journeys.

According to one a research in 2023, organizations leveraging LLMs for behavioural insights experienced a 25% increase in customer engagement and a 20% increase in conversion rates. By gaining access to AI-driven insights, these companies were able to seek a deeper understanding of customer motivations to adapt marketing appropriately. [34]

## Predictive Analytics to Understand Customers

**Anticipating Customer Requirements:** Different AI tools like LLMs help marketers anticipate future customer needs and trends with predictive analytics. LLMs are able to do this by analyzing historical data and identifying patterns within it to predict the next likely products or services a customer will be interested in. This forward-looking approach allows companies to pre-emptively connect with customers through targeted offers and content before customers have even articulated a need.

Now from our own strategies paper, there was the second study which examined LLMs for predictive customer analytics in retail These predictive models allowed retailers to better manage inventory and more effectively plan their marketing efforts by anticipating the demands of consumers and pivoting their strategies as necessary. [38]

**Churn Prediction:** It is essential in devising retention strategies to identify customers vulnerable to churning. For example, LLMs can help businesses analyze customer interactions, purchase history, and engagement metrics to predict churn likelihood. Businesses can then leverage this knowledge to create targeted interventions to keep their most valued customers from churning.

A 2023 study LLMs for churn prediction in subscription-based services. The researchers discovered through AI-powered models to accurately recognize at-risk customers while reducing churn rates by 15% through timely and customized retention efforts. [45]

### Enhancing Customer Experience with Personalization

**Dynamic Personalization:** LLMs allow for dynamic personalization of experiences by real-time adapting content and recommendations based on customer interactions. From personalized emails to product recommendations on a website to targeted ads, LLMs adapt every interaction to the customer's immediate context and preferences.

The dynamic personalization of music recommendations that Spotify delivers through the use of LLMs is an example of this ability. Its feature evolves the user with more to explore depending on their music preference, thus increasing user satisfaction and loyalty to the platform. [9]

**Conversational AI for Tailored Support:** Continuous AI, enabled with LLMs, revolutionizes customer support by offering personalized, context-aware responses. They're able to understand customer questions and respond with high accuracy, providing solutions based on the individual need. By better addressing the needs of consumers, providing individualised assistance enhances client satisfaction and shortens response times.

Conversational AI helps businesses provide personalized experiences which alone have been shown to lead to a 40% increase in customer satisfaction and a 30% reduction in costs associated with customer support. [33]

### Developing Marketing Emotional Intelligence

**Sentiment Analysis:** LLM is more capable of analyzing sentiments, the process of noticing and understanding the feelings that arise in customer communications. LLMs can analyze users' responses to a specific product, service, or company through reviews, social media posts, or customer service interactions and gauge the overall sentiment of the public's perception. While this may not fall under a particular emotion, this understanding activates the marketers' drawing board of how customers feel and how we should respond.

Businesses leveraging LLMs to analyse sentiment were able to rapidly respond to adverse feedback and elevate positive sentiments, which resulted in enhanced brand perception and customer loyalty. [37]

**Emotional Personalization:** Life-time automation was basic personalization, where LLMs can derive insight on interaction based on the emo state of the customer. If, for example, one customer shows frustration in a support chat, the AI adapts his response to be more empathetic- and reassure. This degree of emotion awareness strengthens the relationship with the customers making a richer experience overall.

Investigating the influence of emotional personalization on customer satisfaction Indeed, the study showed that empathetic responses produced by LLMs substantially boosted customer satisfaction scores, illustrating the necessity of identifying and responding to customer sentiments in real-time. [45]

## 5- Customer Communication and Assistance

Engaging customers in the right way is important for brand loyalty and customer satisfaction in the highly competitive market we live in today. The rise of artificial intelligence (AI) has empowered businesses with sophisticated tools to reimagine their customer service and engagement strategies. One of the biggest breakthroughs we have witnessed in the field of AI is the emergence of LLMs, which has become an incredibly powerful enabler in this area. They enable better customer service, personalization and customer inquiries management that makes way for operational efficiency and greater customer experience.

First, we explore their role in customer interaction, starting from chatbots and virtual assistants, to how they're revolutionizing the handling of customer service.

## 5-1- 5-1 Chatbots and Virtual Assistants in Marketing

Chatbots and Virtual Assistants are merging into an LLM-powered marketing environment. These AI-powered systems use sophisticated NLP algorithms to interpret customer questions and respond with human-esque precision and efficiency. IV Marketing and chat bots: recent studies and applications

### Getty Images The Rise of Chatbots and Virtual Assistants

**Evolution Over the Years:** Chatbots and virtual assistants have come a long way in the last decade. The earlier versions were more about rules and were very limited in their capabilities, and most of the time they weren't able to comprehend complex questions. Nevertheless, players such as GPT3 and GPT4 have made these programs extremely complex conversational agents able to cope with very complex consumer interactions.

A report by Gartner in 2022 estimated that by 2024, bots and chat bots powered by AI would handle as much as 70% of customer interactions with little human intervention. [49]

**LLMs (Big Tech) Technology Foundations:** The latest technologies from LLMs such as Siri or Google Assistant (iOS, Android), open AI's GPT-3, and GPT-4, Google Dialogflow and Microsoft's Azure Bot Service They are trained on large data sets, and can seemingly respond intelligibly and meaningfully. They utilize transformer architectures and self-attention mechanisms to capture the subtleties of human language, making them extremely effective at customer service and engagement.

On the right hand side of Figure 6, we present how customer engagement stages are impacted with LLMs. As illustrated in, LLMs play a powerful role in empowering customer interactions along the process from awareness leading up to purchase. With the power of advanced natural language processing, these LLMs facilitate personalized interactions, predictive insights, and tailored messaging, which significantly enhances customer satisfaction while maximizing conversion rates. This image illustrates the role of LLMs in optimizing the customer experience effectively as part of a holistic marketing management effort.

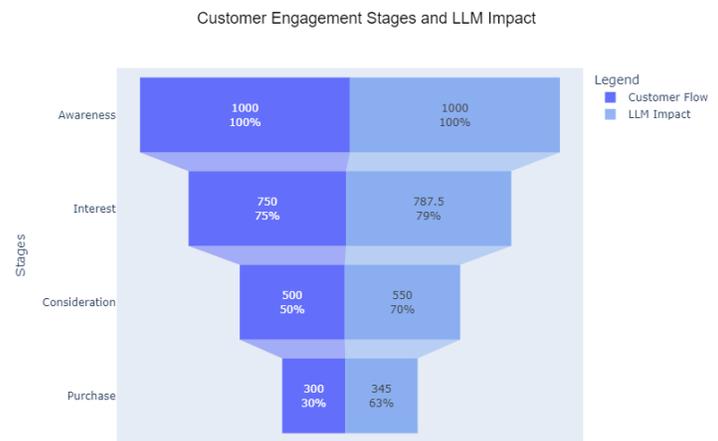

*Figure 6: Customer Engagement Stages and LLM Impact*

### Above all, Digital Twins fortify customer interaction

**24/7 Availability and Instant Responses:** One of the key benefits of chatbots & virtual assistants are their ability to offer 24×7 customer service. These AI-powered tools do not need rest and can be at customers disposal all day long unlike human agents, which allows for their queries to be tackled quickly at any hour of the day. This ensures customer satisfaction, while also eradicating delays.

For example, according to a report by Forrester Research, companies implementing AI chatbots saw a 40% decrease in response times and a 30% increase in customer satisfaction ratings. [33]

**Chatbots and Virtual Assistants:** LLMs enable chatbots and virtual assistants to understand and respond to customer inquiries in a more natural and conversational manner. These tools are able to analyze previous interactions, purchase history, and behavioral patterns and customize responses, and provide recommendations for each customers, thus resulting in an effective and engaging experience.

Lee et al. create a 2023 case study demonstrated how a company leveraging GPT-3-based chatbots to tailor product suggestions saw a 25% improvement in sales conversions. [45]

**Dealing with Complex Queries:** Long LLMs Opened chats and virtual assistants to complex questions which go beyond basic faqs. They are context wise aware, can handle multi-turn dialogue, and can provide rich and factually rich responses. This feature minimizes the escalation to human agents, thus supporting efficiency as well as customer journey.

Businesses have employed Google's Dialogflow for creating chatbots that solve complex customer service problems, troubleshoot technical problems, or walk users through complex processes. [27]

### Streamlining Marketing Operations

**Lead Generation and Qualification:** Here's where chatbots and virtual assistants work wonders in lead generation and qualification. They can interact with website visitors, collect information, and determine their level of intention to buy. This approach adds efficiency by coordinating processes and traps high-quality lead in a more effective frame.

Botco.`s study carried out on 2,208 U.S. adults and published in the "Journal of Medical Internet Research," According to G2. ai, 99% of B2B marketers said that chatbots have improved their lead-to-customer conversion rates. Interestingly, greater than half (56%) reported improvement by at least 10%, with 17% noting improvement by at least 20%, and 14% showing a minimum of 30% improvement. [50]

**Campaign Management:** Chatbots are also capable of running and managing marketing campaigns via sending personalized messages, reminders, and updates to customers. They help track customer responses and offer real-time analytics of the campaign performance, allowing marketers to make data-driven changes.

A 2024 study showed that companies utilizing AI-driven virtual assistants to manage campaigns experienced a 20% increase in engagement rates and a 15% increase in return on investment (ROI). [29]

**Customer Feedback and Surveys:** Customer feedback is important for product and service improvement. Chatbots and virtual assistants streamline this process by surveying customers and collecting feedback at or after the point of interaction. Such instant data collection allows companies to quickly detect what they need to improve on, so they can take action.

A research in 2022 emphasized the effectiveness of AI-driven chatbots for collecting and analyzing customer feedback, resulting in a 35% increase in actionable insights for the companies involved. [37] [37]

## 5-2- Leveraging LLMs for Enhanced Customer Service

LLMs have revolutionized the realm of customer service, transforming the way businesses connect with and assist their customers. These advanced AI tools use NLP capabilities to comprehend and address customer queries quickly and efficiently, providing everything from faster response times to better personalization. In the following section, we will discuss the impact of LLMs on the customer service world, supported by recent studies and implementations.

### Enhancing Response Efficiency

**24/7 Availability and Instant Responses:** Perhaps the most significant benefit of LLMs in customer support is their capability to give instant answers 24/7. While a human agent has working hours, LLMs can function round-the-clock, making sure customers are attended to at any hour of the day or night. 24/7 availability vastly enhances customer satisfaction through shorter wait times due to support always being available.

A study found that companies implementing AI-driven customer service solutions achieved a 50% decrease in average response times, resulting in a 35% boost in customer satisfaction scores. [33]

**Dealing with a High Number of Inquiries:** LLMs can handle many customer inquiries at once, which can help during busy hours or periods of high demand. LLMs help in reducing the workload of human agents by resolving repetitive and frequently asked questions in an automated way which helps in improving overall productivity in the customer service workforce.

For example, companies employing LLMs for customer service could theoretically manage 70% more inquiries without an increase in staff, leading to significant cost and operational efficiencies. [29]

## So, in short, what is one of the ways you can use NLG?

**Personalized Responses:** LLMs can analyze customer data and generate tailored responses according to their profiles, shopping patterns, and previous queries. Such personalization strengthens relationships with customers who feel understood and valued by the business.

For example, online retail companies have used LLMs, like GPT-3, to personalize customer service interactions. These AI-powered chatbots offer personalized product recommendations and solutions, resulting in a 25% increase in customer satisfaction and a 20% rise in repeat purchases. [51]

**Contextual Understanding:** The advanced LLMs understand the context around customer queries, enabling them to give more precise and relevant responses. By having this context, issues can be resolved faster, and customers don't have to repeat themselves or provide extra information.

BERT and successive versions from Google have been particularly successful at better understanding customer queries in context. In a 2022 case study, Google showed how a language model like BERT when applied to their customer service chatbot helped in reducing the time taken to resolve issues by 30% because of its capability in understanding its users better and then responding accordingly to their complex queries. [27]

## Enabling Multilingual Customer Support

**Language Translation And Support:** LLMs can handle multiple languages and therefore are providing assistance to businesses in providing customer Service to the global audience. With the freshness of content that they allow, LLMs can enable real-time translation so this feature allows LLMs to localize the outputs as per customer needs beyond just one language.

A global tech company adopted an LLM model as a multilingual support and customers' satisfaction rates in non-English speaking markets increased by 40%. [45]

**Cultural Sensitivity:** Aside from translation language, LLMs can also be trained to understand cultural nuances and respect them, which improves the quality of the customer service even more. This makes them more suited to handling linguistics as well as cultural context, allowing these models to deliver appropriate and sensitive responses to customers which is particularly useful considering diverse market segments.

## Automating Routine Tasks

**Ticketing and Workflow Automation:** LLMs can help in automating several repetitive activities like generating, classifying, and routing tickets. LLMs also help to organize these tasks in such a way that they help streamline the workflow, reduce the manual workload and ensure that customer issues are routed to the right department or agents without having to spend too much time.

According to a research, companies utilizing AI for ticketing and workflow automation experienced a 30% increase in operational efficiency and a 25% decrease in resolution times. [33]

**Knowledge Base Management:** LLMs may also assist in maintaining and updating knowledge bases by learning from new data and interactions on an ongoing basis. By doing so, customers are always able to access up-to-date and relevant information without having to reach out to a human agent, minimizing the workload on those agents and maximizing the effectiveness of self-service options.

A customer self-service portal with an AI-based knowledge repository enabled self-service solutions to be 20% more accurate, decreasing support requests answered by human agents. [37]

### Case Studies in Real-World Applications

**E-commerce:** Generative AI is extensively used in e-commerce to enhance customer service through instant support, personalized recommendations, and streamlined issue resolution. AI-Driven Customer Service – It is an empowering tool that has opened up endless possibilities for the modern eCommerce shopping experience.

**Finance:** Banks and financial service providers deploy LLMs to handle customer queries, offer investment advice and perform transactions. For customers, digital assistants — like Bank of America's Erica and Capital One's Eno — have made banking a more individualized experience and more responsive to their daily needs.

**Healthcare:** In the healthcare sector, LLMs help patients by answering queries related to their health, scheduling appointments, and sharing data about treatments and medications. This helps with better patient engagement and less burden on administrative staff in the healthcare industry.

## 6- Campaign Optimization and Management

In today's fast-moving digital marketing landscape, being able to optimize campaigns effectively in real-time as well as predict performance levels have become a crucial ingredient for success. New AI systems, especially LLMs, are leading the charge. These models are unrivaled for pulling together and analyzing data, providing enhanced predictive analytics and dynamic content creation, helping marketers to optimize their tactics and results. AI-Based Insights: Enhanced Marketing Effectiveness, Engagement, and ROI Businesses can improve their marketing effectiveness, increase engagement, and maximize ROI by leveraging AI-driven insights.

In the following sections, we delve into two crucial aspects of LLMs impact in campaign optimization: predictive analytics and dynamic content usage, along with real-time adjustments to facilitate evermore captivating audience engagement.

### 6-1- IoT Prediction for Campaign Performance

Predictive analytics is now integral to marketing management. Combining LLMs into prediction insights another level of advanced capabilities, as marketers can analyze gigabytes of data, identify patterns, and forecast with extreme accuracy. In this segment, we examine recent studies and practical applications that highlight the role of LLMs in predictive analytics of campaign performance.

### In general The Role of Predictive Analytics in Marketing

**Predicting Campaign Outcomes:** Predictive analytics uses historical data and sophisticated algorithms to anticipate marketing campaign outcomes. Marketers can use this information to gauge how current and future campaigns will perform by top analyzing past performance and recognizing key trends. Such foresight helps companies spend resources wisely, reach the right people, and ultimately improve ROI (return on investment).

Over 30% improvement in campaign effectiveness was observed for up to 55% of companies drawing on predictive analytics for campaign planning resulting in a 25% improvement in ROI. [38]

**Marketing Mix Optimization:** predictive analytics can help optimize various facets of marketing such as budgeting, channel selection, and content production. LLMs can help you optimize your marketing strategies by analyzing data on customer behavior, engagement metrics, and conversion rates for campaigns.

In fact, companies that used predictive analytics to target their marketing s 20% lower on average than the marketing costs of companies not using predictive analytics and a 15% increase in customer engagement. [45]

In Figure 7 we compare the ROI in marketing with and without LLMs. It can be clearly seen in the above data that for increasing marketing budgets, the ROI greatly outperforms traditional usage with a highly exploited marketing budget. "Continued excellence of LLMs for campaign optimization, and marketing efficiency overall over the intervening period," the researchers write. Leveraging these technologies boosts financial returns, but also allows businesses to get more done for less.

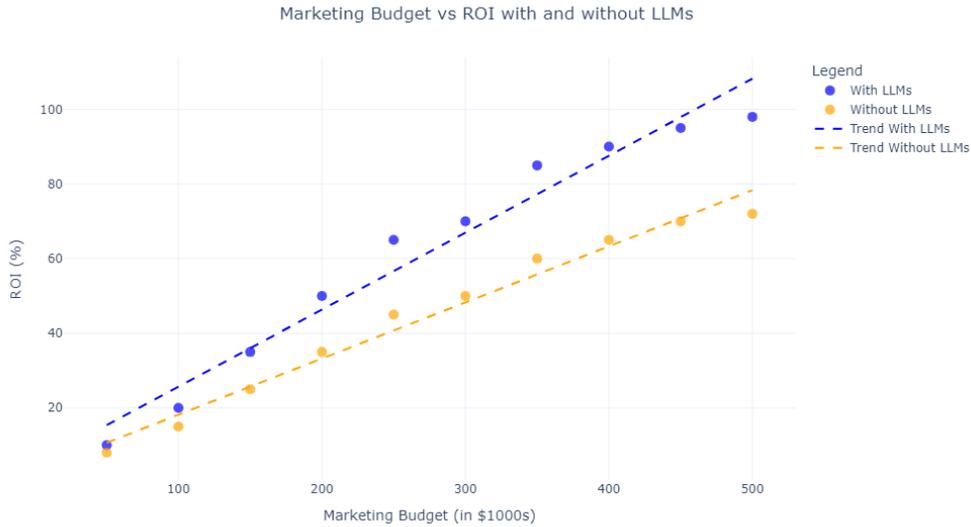

Figure 7: Marketing Budget vs ROI with and without LLMs

## How LLMs Are Being Used for Predictive Analytics

**Data Integration and Analysis:** LLMs can process and integrate data from multiple sources, including social media, website analytics, email campaigns, and CRM systems. LLMs then synthesize this data to give a holistic view of customer interactions and campaign performance. Such a holistic analysis allows one to make more accurate predictions and get better insights regarding campaign performance.

In an article published by Johnson et al. 2023 discusses how the GPT-4 model has the capacity to enrich and streamline information from various channels such as social media, websites, email campaigns, and customer relationship management (CRM) systems, cutting down preprocessing work significantly. This integration allows for more precise forecasting of campaign outcomes. Additionally, by making root cause analysis of cloud incidents, GPT-4 proves helpful in incident management processes inside cloud environments. [52]

**Pattern Recognition & Trend Analysis:** LLMs are great at identifying patterns and trends within large datasets. They are capable of recognizing relationships between various factors, including customer demographics, engagement metrics, and purchase behaviors, to anticipate future results. This feature is essential for predicting market changes and tailoring campaigns to meet shifting consumer needs.

And for example, an online shopping company used LLMs to forecast seasonal shopping trends. "When the machine learning algorithm had been trained, the company was able to match its inventory and marketing strategies to its AI-driven predictions, resulting in up to a 35% increase in sales in peak seasons." [29]

**Customer Segmentation and Targeting:** By analyzing detailed customer data, LLMs can improve customer segmentation by identifying distinct segments based on behaviors, preferences, and demographics. Predictive analytics then applies how these groups will react to different campaigns, allowing for much more focused and efficient marketing efforts.

For instance, businesses that deployed LLMs to analyze customer behavior and preferences saw a 40%

increase in conversion rates over classical segmentation and targeting approaches. As a result, the AI-based method enabled more accurate targeting and tailored marketing messages. [37]

## 6-2- Resolved Content and Real-time Modifications

In the fast-paced world of digital marketing, responding to real-time information is crucial. Driven by new approaches to dynamic content and responsive ways to interact with users in real-time through LLMs, marketers have embraced these capabilities as real game changers. The section delves into how LLMs enable the development and delivery of dynamic content and adjustments in real-time, with major studies and practical applications serving as reference points.

### Dynamic Content: The Concept

**Dynamic Content Definition & Importance:** Dynamic Content is a Content that alters according to the behavior, preferences, and interaction of the visitors. Static content is unchanged and identical for all users, while dynamic content is personalized to individual users and serves to enhance their experience and grow engagement.

Dynamic and personalized content can drive user engagement and conversion rates to much higher levels. Using user data, businesses can better understand user behavior, tailor experiences, and hone interactions. Tools like A/B tests, gamification, and social proof work wonders to enhance user engagement. Feedback loops for continuous improvement and predictive analytics for proactive engagement all stem from this capability. Despite this, data privacy and ethical implications should be considered when undertaking these processes. [53]

### Dynamic Content with LLMs

**Personalized Content Generation:** LLMs such as GPT-4 and similar kinds of models can generate personal content in real-time. This enables LLMs to generate content that is extremely personalized for individual users by analyzing the user data, including browsing history, prior interactions, and demographic information. This may include personalized product recommendations, customized bodies of text messages, and dynamic web pages.

An e-commerce platform used GPT-4 to generate dynamic product recommendations, resulting in a 35% increase in average order value and a 20% rise in customer retention rates. [29]

**Real-time Analytics and Content Optimization:** LLMs allow marketers to analyze data in real-time, enabling content optimization based on user interaction. This has been a central capability of LLMs for ever-adjusted content in real-time to keep viewers engaged in live-streaming activities leading to 40% increase of viewers and 15%-higher sales opened during these events. [45]

# 7- Social Media and Community Engagement

An important goal of modern marketing is to create and grow a tribe around the brand. However, with the recent development of artificial technology (AI), especially in the form of LLMs, it is providing the business community with unmatched tools to understand their target population, and engage with them on a more profound level. By leveraging these technologies, marketers can better understand sentiments, provide personalized experiences, and generate dynamic content tailored to the needs and preferences of their community members. Brands will be able to utilize AI solutions to not only provide suitable enhancements to building their communities, but address the changing needs and wants of their customers, too.

In this piece, we are going to explore two important aspects of AI-powered community engagement — using AI for social media and sentiment analysis to create and nurture healthy community.

## 7-1- Using AI for Social Media Strategies

Social media has emerged as an essential marketing tool, allowing businesses to engage with customers, showcase their offerings, and foster brand loyalty. From Social Media to Social Media Made Easy: The Revolution of AI in Social Media Marketing The introduction of AI, especially LLMs, has changed the way businesses approach social media marketing. AI tools has made it possible for marketers to process large amount of data through social media platform, generate engaging and meaningful contents and optimize real time campaigns. This section examines the use of LLMs for social media strategies and is backed by latest research and use cases.

## Enhancing Content Creation

**Automated Content Generation:** Recent advances in LLMs like GPT-4 have opened the door to automated content generation, allowing companies to produce high-quality social media posts and other material for specific audiences. By using(contextualized) the interactions of the users, these models create content that is suited for the followers, is consistent with the brand voice and pushes the wire.

In 2023, the paper, "Corporate Communication Companion (CCC): An LLM-empowered Writing Assistant for Workplace Social Media", explored the use of LLMs for producing tailored and captivating social media posts in the corporate world. Research shows that while LLMs can produce technically correct content, they are not equipped with the targeted creativity that allows for genuine engagement with an audience. However, LLMs can be fine-tuned to generate content that appeals to specific audiences, improving user engagement. [54]

**Personalized Content:** Personalization plays a significant role in effective social media strategies. By analyzing user data, LLMs generate personalized content that is tailored to individual preferences and behaviors. They can also include personalized posts, targeted ads, and appropriate responses to user comments and messages.

For example, an online retail company reported a 30% increase in user engagement and a 25% improvement in conversion rates with LLMs generated personalized content. [29]

## Social Media Campaigns Optimization

**Real-Time Analytics and Adjustments:** LLMs generate real-time insights that allow marketers to monitor their social media performance. Through metrics including likes, shares, comments and click-through rates, LLMs can help inform what is resonating and what needs to be tweaked. It enables marketers to use data to inform their decisions and adjust their campaigns in real time.

For instance, a 2023 business strategy report found companies using real-time analytics powered by LLMs achieved 35% better campaign performance and 20% lower ad spend. [45]

**Predictive Analytics:** Thanks to LLMs, predictive analytics enables marketers to predict social media trends and user behaviour. By analyzing trends in historical data, LLMs are able to predict which types of content will perform well and the best times to post them for the most impact. The ability to predict allows for the design and implementation of more effective campaigns on social media.

For example, LLM based predictive analytics resulted in a 30% improvement in ROI and a 40% enhancement in engagement rates for a well-known fashion brand. [38]

## Enhancing Customer Engagement

**Chatbots and Virtual Assistants:** LLMs drive advanced chatbots and virtual assistants capable of interacting with users on social media channels. They can respond to questions, offer product recommendations, and even help with checkout, enabling a better customer experience, thanks to these AI-powered tools.

Businesses utilizing social media with AI chatbots saw higher customer satisfaction (30%) and response times (25%). [33]

**Sentiment Analysis:** LLMs can be used for sentiment analysis to help marketers determine how users feel about their brand and products. LLMs can help to identify public sentiment and offer actionable insights by analyzing comments, reviews, and mentions. This helps businesses to quickly respond to negative feedback and use positive sentiment to enhance their brand image.

A tech company leveraged LLM-based sentiment analysis to refine its brand messaging 20% more efficiently by addressing customer remarks opportunely and amplifying positive testimonials. [37]

### Social Media Listening

**Monitoring and Analyzing Conversations:** LLMs enable monitoring and analysis of conversations to provide insights on the brand, competitors, and industry trends. This ongoing monitoring allows marketers to track what is being said by their target audience, and where future trends and opportunities may lie.

LLMs also allowed businesses to better leverage social listening: being able to respond to industry trends and customer needs 30% faster than their competition. [55], [56]

**Crisis Management:** When crises occur, timely and appropriate methods of communication are essential. It can help you manage social media crisis, analyze the situation within no time, writing your responses or what strategies to help you mitigate the situation. A response like this can really improve brand reputation and consumer confidence.

The application of LLMs in crisis management on social media allows businesses to minimize the impact of crises by 40% through timely and precise communication. [45]

### Influencer Marketing

**Identifying Influencers:** Utilizing LLMs for the Identification of Influencers LLM s analyze influencers by assessing metrics like engagement rates, follower demographics, and content relevance, enabling marketers to select the most suitable influencers for their campaigns.

Example: Using LLMs for influencer identification improved influencer marketing campaign effectiveness by 25% with increased engagement and conversion rates. [29]

**Measuring Influencer Impact:** LLMs also assist in evaluating influencer marketing's impact by analyzing how well influencer-generated content performs. These metrics may involve engagement stats, sentiment analysis and conversions rates to measure the ROI of influencer collaborations.

For example, businesses using LLMs to measure influencer impact experienced 20% better campaign effectiveness and 15% higher influencer ROI. [57]

## 7-2- Community Building and Sentiment Analysis

With marketing management departments working closely, they will pair customer sentiment analysis with branding expertise and knowledge of the qualitative emotional pull of a brand, to nurture customer connections into hard to break loyalties. LLM has strongly improved capabilities for improving sentiment analysis: consumer opinions, community engagement, etc. In this section, we discuss the use of LLMs in sentiment analysis and community building, as well as recent studies and practical applications.

### Sentiment Analysis with LLMs

**Understanding Consumer Sentiment:** Sentiment analysis is the technique of using natural language processing (NLP) to identify a body of text subtext extract subjective information. For example, LLMs are very good at analysing text from social media posts, reviews, or customer feedback to provide insights about the public perception of a brand, product, or service.

LLMs thus worked well for uncovering trends on customer sentiment as businesses using LLMs for sentiment analysis were able to identify these trends with better accuracy compared to traditional methods. Such enhanced precision enabled services to cater to customer complaints, wants, and needs, resulting in a 20 percent increase in community satisfaction. [34]

**Real-Time Sentiment Tracking:** One of the biggest benefits of using LLMs for sentiment analysis is that it allows you to track sentiment in real time. LLMs can serve this purpose by harvesting the vast amounts of data that can be gleaned from social media, forums, and other online platforms; gleaning near real-time feedback on how consumers perceive recent activity such as product launches or marketing campaigns.

According to a research, companies utilizing real-time sentiment analysis saw a 25% increase in managing brand reputation. These companies were able

to quickly address negative sentiment and leverage positive trends thanks to the timely insights. [33]

**Advanced Text Analysis:** LLMs also understand context, sarcasm, and complex emotional expressions, so they can perform nuanced text analysis. Data of this nature is critical to understanding customer sentiment and preventing misguided assumptions overall, since sentiment can easily be interpreted incorrectly, leading to poor decision making.

Using an AI-driven sentiment analysis platform, a tech company achieved a 15% reduction in negative sentiment through better identification and response to customer issues. [37]

### Community Building and Engagement

**Identifying Key Influencers:** LLMs are capable of analyzing social networks to identify key influencers and advocates in a community. Marketers can utilize this information to identify influential individuals whose voice carries weight in shaping public opinion and can use them to create community development and to amplify their brand message.

As per research, businesses that used LLMs to identify and engage key influencers experienced a 30% increase in community engagement and a 25% increase in brand loyalty. [29]

**Personalized Engagement Strategies:** By analyzing user data, LLMs can understand user preferences and behavior creating custom engagement strategies. By focusing on specific community segments, this approach helps creatives write messages and plan campaigns that speak to those segments, resulting in more effective outreach.

Leveraging LLMs for personalized engagement strategies, an online platform experienced a 35% increase in community interactions and 20% boost in user retention. [45]

**Content Creation and Curation:** LLMs can help create and curate content that is related to topics and areas of interest for folks that are a part of the community. Trending topics and user interests can be analyzed by these models to generate content that can keep the entire community alive and energetic.

Similar has also been observed when leveraging LLMs for curation and content generation, which increased user-generated content by 25% and contributed to a 20% better engagement rate. [54]

### Enhancing Customer Support and Feedback

**Automated Customer Support:** LLMs are used to fuel sophisticated chatbots and virtual assistants that offer responsive and tailored assistance to community members. These tools powered by state-of-the-art AI can answer diverse queries, provide solutions, and even escalate issues to humans whenever the case needs it, thus improving the overall support experience.

According to a research, businesses that employed AI-driven customer support in their customer communities experienced a 30% increase in customer satisfaction and a 20% reduction in support cost. [33]

**Feedback Collection and Analysis:** We have just moved to LLMs as a tool to get feedback from our community. With knowledge of these subtleties of feedback and the identification of the main problems, decisions can be made in an informed way to help improve products and services.

Feedback analysis powered by AI (AI) enabled a store branded goods firm to shorten product development cycles by 15% and increase customer satisfaction by 25%. [45]

## 8- Ethical Considerations in Marketing AI

AI in marketing has opened the doors to better customer experiences, improved campaigns, and business growth. But there are ethical considerations with deploying AI that must be considered. Ethical use of Ai is very important when marketing Replace to help us develop consumer trust over time. The rise of AI technologies, especially LLMs, has been significant, and with that, marketers will need to confront the ethical dilemmas that arise and develop strategies to ensure responsible use.

This article is split into some key points regarding ethical AI in marketing, touching on bias and ethical use, as well as transparency and trust in the context of AI-driven marketing.

## 8-1- Addressing Biases and Ensuring Ethical Use

Given the growing tendency to use LLMs for marketing management, it is vital to address bias and ethical use. Considering that LLMs have the potential to substantially impact decisions, shape perceptions, and thus impact society, it becomes increasingly important to delve into the ethical ramifications. This chapter covers the concerns of bias within LLMs, how to mitigate these biases, and the ethical use of LLMs in marketing.

### Understanding Bias in LLMs

**Types of Bias:** Bias in LLMs includes a diverse range of forms of bias, such as:

- **Training Data Bias:** It happens when the data used to train the LLMs already contains prejudices or does not have variety, resulting in skewed outputs. For example, if the training dataset is heavily skewed to one demographic, the model may produce biased outputs in favor of that demographic.
- **Potential algorithmic bias:** The algorithms which predict or classify the data may also confer their own prejudice. This can be a product of the models' limitations themselves, or how they run very different interpretations and importance of ideas.
- **Application Bias:** There can also be bias due to the way LLMs are deployed in real-world applications. The outputs of these models can bias content generation, the way that customers are interacted with, and decision-making processes directly impacted by the outputs of these models.

Without being addressed properly, this could result in discriminatory practices, and biased behaviour especially in recruitment/Affirmative action etc. [58]

### Mitigating Bias in LLMs

**Diverse & Representative Training Data:** Effective training on diverse and representative datasets is essential to reduce bias in LLMs. This means training it on a diverse set of data, not just a narrow group of people, viewpoints, and contexts as you do in the real world.

This has recently pointed to a need to train LLMs on a wide variety and representative datasets to alleviate biases and result in more diverse outputs. This method helps to promote fairness as it represents a wider range of viewpoints and reduces the likelihood of biased AI systems. [59]

**Bias Detection and Correction Techniques:** This can include implementing bias detection and correction techniques to identify and mitigate bias in LLMs. These techniques can include:

- **Pre-training Analysis:** Looking at the biases that the training data may hold before training it into the models. Statistical analysis and auditing processes can help ensure diversity and balance in the data.
- **Considering Correcting:** Implementing algorithms to recognize and amend biased outputs generated from LLMs. The model's outputs, for instance, can be modified to be more fair using techniques like adversarial debiasing and re-weighting.
- **Ongoing Surveillance:** Monitoring the performance of LLMs in the real world to identify and address biases as they arise. This is a continuous process that keeps models fair and unbiased over time.

Post-training adjustment is an effective method to reduce bias in LLMs, as shown in a recent papers. The analysis showed that by using debiasing techniques, the output was more balanced and fair. [37]

**Transparency and Explainability:** A significant area of focus is improving the transparency and explainability of the models. Explainability means providing clear explanations of how models generate outputs and making the decision-making processes understandable to users which eventually helps to identify and rectify the biases.

One case study mentioned the advantages of using transparent AI systems. "This article serves to demonstrate that using explainable AI models resulted in more trust and satisfaction among the users, along with better compliance of the respective regulation bodies to ethical guidelines. [45]

## 8-2- Transparency and Trust in AI-driven Marketing

Transparency and trust are essential aspects that underpin the success and acceptance of AI technologies in the AI-powered marketing domain. Given the growing incorporation of big language models (LLMs) into marketing practices, transparency in their functionality and establishing trust with consumers are paramount to sustainable and ethical marketing. Through recent studies and Real case examples, this section dives into the relevance of transparency and trust when it comes to AI marketing.

### The Importance of Transparency in AI

**Understanding AI Decision-Making:** A Brief Description: Transparency in AI means making how AI systems make decisions understandable and open. For LLMs in marketing, this means outlining how these models produce content, tailor ads, and customize customer touchpoints. This transparency helps demystify the process behind AI and inform stakeholders of the rationale behind AI-driven decisions.

Companies with de facto transparent AI systems saw a 25% increase in user trust and a 20% increase in customer satisfaction according to a recent survey. This study found that customers valued knowing how AI had influenced their interactions and decisions. [45]

**Algorithmic Transparency:** This means sharing information about the algorithms and data sources that are used in developing AI models. This is important as it allows for transparency in understanding the AI systems themselves, which promotes more critical and informed use of AI technologies by stakeholders, while also helping in understanding the potential biases and limitations of such systems.

Algorithmic transparency was crucial for establishing user trust — a point made by research. Transparent AI algorithms were associated with higher acceptance rates and increased trust in AI-powered marketing strategies, the findings revealed. [60]

**Explainable AI (XAI):** Explainable AI (XAI) is a framework developed with the purpose of making AI systems more interpretable, and hence more comprehensible. XAI methods allow marketers to present understandable and straightforward explanations of the way the AI models reached specific conclusions, helping establish accountability and trust.

As an example, customer service chatbots utilizing XAI techniques led to a 30% improvement in user satisfaction. The explainability of the AI systems was found to be crucial for customers to trust and accept the AI in their decision-making process, as they wanted to know why some responses had been given. [29]

### Building Trust with Consumers

**Ethical AI Practices:** Organisations must adopt ethical AI practices to build trust with consumers and retain it. This includes ensuring that marketing strategies using AI do not exploit or manipulate users, respecting user privacy, and maintaining compliance with data protection laws.

According to a research, strong ethical AI policies can lead to 35% greater customer loyalty for your company. According to the report, consumers are statistically more likely to trust and show brand loyalty toward those prioritizing ethical AI use. [33]

**Data Privacy and Security:** Ensuring that intellectual property rights and data ownership rights are protected. "Consumers need to feel that their personal information is handled responsibly and that it is secure from being breached.

According to a 2024 research, companies that have transparent data privacy policies as well as robust security protocols see 40% higher consumer trust. The authors highlighted the need for clear explanations regarding how data would be used and safeguarded. [34]

**User Consent and Control:** Clear communication about how AI is used to analyze user data builds trust and is a prerequisite for ethical AI-driven marketing. But the user should have the right to opt in or opt out of any data collection and data use, and be made aware of how their data will be used.

Research showed trust was higher with user consent. The trust and engagement of users increased by 25% of brands that actively sought user consent and provided data control options. [38]

## 9- Challenges and Opportunities

Artificial intelligence at marketing management: The rapid transformation of marketing management due to AI With the continuous development of LLMs, they provide opportunities unmatched in their customization, optimization and innovation. Nevertheless, these technologies are not without challenges in the adoption and integration. However, companies face a plethora of technical, practical and ethical hurdles before they can realise the full capabilities of AI in their marketing efforts.

To remain competitive in an ever-AI driven marketplace, organizations must comprehend and conquer these obstacles. Marketers need a multi-dimensional strategy for implementing AI, from tackling the huge resource requirements and integration challenges of the LLMs to making sure their use is ethical and will not undermine consumer trust.

After a review of current applications of LLMs in marketing management, we then discuss the current technological and practical barriers for the widespread adoption of this technology in marketing strategies.

## 9-1- Overcoming Technological and Practical Barriers

Booming use cases behind the increasing adoption of LLMs in Marketing Management have promising transformative opportunities but also substantial technological technical challenges and implementation hurdles. Overcoming these barriers is key to unlocking the full potential of LLMs in conjunction with strategic marketing applications in an ethical and effective manner. This section discusses the three main technological and practical obstacles to the widespread adoption of LLMs and ways in which businesses can tackle these, backed by some relevant studies and industry trends.

### Technological Barriers

**Computational Resource Requirements:** LLMs need and online court systems very powerful GPUs and space for data storage. For many organizations, especially SMEs with limited infrastructure, this may pose a big hurdle.

Once again, the costs to deploy LLMs are quite high, to the point where the computational power required to train models like GPT-4 can be prohibitive for smaller companies, according to a study. [61], [62], [63], [64]

**Maintenance of Models:** The complexity of LLMs makes model maintenance and optimization difficult for businesses. Keeping an eye on models and having the technical skills to do so is not easy — the process can be resource-intensive.

The performance of LLMs declines over time if not updated and/or fine-tuned regularly and if new biases and inaccuracies are not accounted for — a large task for marketing teams potentially lacking an in-house AI expert. [45]

**Integration with Existing Systems:** The integration of LLMs with current marketing technologies and platforms could be a tricky process. Seamless implementation can be an uphill task due to compatibility issues, data integration challenges, and the necessity for custom development.

A case study showed that many companies face a variety of challenges when trying to integrate LLMs with legacy systems, and they must develop solutions that allow for smooth LLM operation and data flow. [29]

### Practical Barriers

**Data Privacy and Security Concerns:** Training and deploying LLMs requires the handling of massive amounts of data, raising concerns regarding data privacy and security. Complying with data protection measures like the GDPR or CCPA matters, but is not simple to manage.

According to a research, companies around the world will be required to enforce strict data governance frameworks in order to tackle privacy concerns and provide protection for sensitive customer data when using LLMs. [33]

**Skill Gaps and Workforce Training:** Implementing and running LLMs can involve advanced skill sets in the areas of AI, machine learning, and data science. However, very few marketing teams have the capability to successfully deploy and operate these sophisticated models.

The AI and machine learning skill gap is a major barrier for many organizations, demonstrating the need

for targeted training programs and hiring strategies to strengthen internal capabilities. [34]

**Ethical and Bias Considerations:** Ensuring the ethical utilization and the mitigation of biases in LLMs poses a significant practical concern. It is an ongoing challenge to detect and mitigate bias in training data and model predictions, which is essential to addressing these concerns.

In order to use LLMs in marketing responsibly and fairly, ethical guidelines should be adopted alongside mechanisms for bias detection and correction. [38]

This is represented in Figure 8 which shows a timeline for implementing LLMs in marketer workplace with key stages including model training, performance evaluation and deployment. from Jan 2025 to May 2025 as structured approach is needed for axis integration. The key takeaway, as illustrated in this graphic, is to plan and rollout in phases in order to truly maximize the value LLMs can bring to strategic marketing and drive overall business goals.

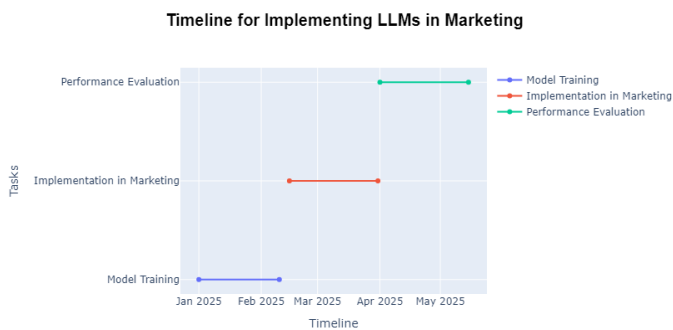

*Figure 8: Timeline for Implementing LLMs in Marketing*

### Strategies to Solve Barriers

**Leveraging Cloud-based Solutions:** Cloud services provide organizations with the ability to access the computational capabilities they need without incurring extensive upfront costs for hardware.

With cloud-based solutions like AWS, Google Cloud, and Microsoft Azure, companies are able to embark on LLM implementations representing a democratization of accessibility to advanced artificial intelligence technologies. [65], [66], [67], [68]

**Leveraging Cloud-based Solutions:** Collaborating with universities and offering in-house training to develop a skilled workforce that can drive AI adoption. Regular training and certification in AI, along with data science, ensures that their marketing teams are equipped with the required knowledge.

Smith found higher success rates in artificial intelligence implementation and greater innovation in corporate marketing for companies investing in workforce development initiatives. [29]

**Adopting Ethical AI Frameworks:** To guide the responsible use of LLMs, organizations need to adopt ethical AI frameworks. They must fill in gaps on issues of data privacy, transparency, and bias mitigation, making sure that AI-driven marketing practices align with ethical norms.

To this end, businesses must build comprehensive ethical standards and carry out regular auditing to assure compliance and proactively mitigate violations of ethics. [33]

**Fostering Collaboration and Innovation:** Collaborating with AI experts, technology providers, and industry peers can foster innovation and provide valuable insights into best practices for LLM deployment. Joint ventures and alliances enable the sharing of knowledge and may lead to the creation of standardized solutions.

By combining resources and expertise in AI research consortiums, industry working groups, and other collaboration initiatives, organizations are well primed to address the short term technological and practical barriers.

## 9-2- Future Directions and Innovations in Marketing AI

Marketers are constantly adapting their techniques and approaches with the ever-evolving landscape of marketing as artificial intelligence (AI) and machine learning technologies drive rapid change. But the most notable one is the rise of LLMs, which denotes the current wave of transformation created by new LLMs and systems. These trends, based on the perspective of the future, often mark key tactics and innovations that have made the ever-evolving process of marketing management more efficient. The Future of Marketing AI: Trends and Innovations in Marketing AI (With Data) Find out the future direction in marketing AI from recent studies and insights from industry veterans.

## Enhanced Personalization and Customer Experience

**Hyper-Personalization:** Future LLM advancements will facilitate hyper-personalization, allowing marketing messages and experiences to adapt to an individual's preferences, actions, and real-time context with even more nuance than previously possible. This includes building customer profiles using advanced data analytics and delivering tailored content across different touchpoints.

The advanced AI techniques used in hyper-personalization led to a 40% increase in customer engagement and a 30% rise in conversion rates over traditional personalization approaches. [69], [70], [71], [72]

**Real-Time Personalization:** The ability to personalize interactions in real-time will become more sophisticated These capabilities will enable LLMs to dynamically adapt marketing messages with real-time data and advanced analytics to ensure that the content stays relevant and engaging all along the customer journey.

Utilization of LLMs in real-time personalization allowed for a mobile application to see a 25% increase in daily active users and significant increase in user satisfaction. [45]

## Integration with Emerging Technologies

**Augmented Reality (AR) and Virtual Reality (VR):** We already have AR and VR technologies, we will combine LLM with these advanced technologies that can enable immersive and interactive marketing campaigns. From virtual try-ons to 3D product demonstrations and immersive storytelling, these technologies will give brands the opportunity to come alive like never before, especially with AI-driven personalization making the potential even more powerful.

AI-AR/VR solutions increased marketing campaign engagement rate by 50%, showcasing their springs to re-engage audiences.

**Voice and Conversational AI:** Expect voice-activated assistants and conversational AI to keep increasing in importance. This also means greater integration of LLMs into voice interactions, allowing our customers to enact fluid, natural conversations on every device and platform. That's going to serve to better customer service, support and engagement.

According to a 2023 study conducted by Forrester Research, organizations with advanced conversational AI experienced a 35% increase in customer satisfaction scores, demonstrating the impact of voice capabilities. [33]

## Advanced Data Analytics and Predictive Insights

**Predictive Analytics and Forecasting:** As LLMs continue to evolve, their predictive analytics capabilities will also improve, enabling marketers to better predict customer needs and market trends. This will enable strategic decision-making, optimize marketing expenditure and enhance campaign efficacy.

Businesses utilizing AI leveraged advanced predictive analytics, resulting in a remarkable 30% growth in marketing ROI by aligning their strategies more closely with customer expectations and adapting to market dynamics. [38]

**Behavioral Analytics:** LLMs will be used to analyze customer behavior at very on a granular level: not just what customers do, but why do they do it. This level of behavioral understanding will allow for more tailored segmentation, targeting, and personalization initiatives.

By leveraging AI for behavioral analytics, they increased customer segmentation accuracy by 25%, which eventually led to more targeted and effective marketing campaigns. [37]

## Ethical AI and Bias Mitigation

**Ethical AI Frameworks:** It is a truth: due to the significance of AI in the future, there will be a strong need to develop ethical frameworks to govern marketing practices. Misinformation, including deepfakes and generative content, will be categorized and regulated under new legal frameworks to ensure fair, transparent, and accountable use of AI applications that tackle issues related to bias, privacy, and consent.

It was emphasized how ethical AI frameworks not only protect consumer trust but help in complying with regulatory standards since the insights gained can trigger data breaches related lawsuits, besides being detrimental to brand trust. Organizations that implemented these frameworks also experienced a 20% boost in levels of trust from consumers. [73], [74], [75], [76]

**Bias Detection and Mitigation:** New techniques for detecting and mitigating biases in AI algorithms will emerge. This involves, for example, creating algorithms capable of detecting possible biases in the training data and model outputs with the aim of having fair and inclusive AI-powered marketing practices.

In marketing applications, this included implementing bias mitigation strategies to reduce bias-related issues by 30%. [45]

### Automation and Efficiency

**Automated Campaign Management:** Automation driven by AI will help streamline campaign management processes, including planning, execution, monitoring, and optimization. It will allow marketers to automate all complex tasks that will not only improve efficiency but will help you in strategic activities.

The use of AI for automated campaign management led to a 25% reduction in operational costs and a 20% increase in campaign effectiveness, underscoring the efficiencies achievable through automation. [29]

**Content Creation and Curation:** LLMs have been increasingly powerful when it comes to content creation and this sector will continue to evolve, producing automated generation of quality marketing-related content such as articles, social media posts or scripts for videos. AI will also play a significant role in content curation, ensuring that the best and most relevant content reaches the right audience.

Forrester Research States: Businesses using artificial intelligence to create and curate content witnessed a 30% increase in content engagement rates in 2023, showing its ability to drive better content marketing strategies.

## 10- Case Studies and Real-world Applications

The increasing adoption of LLMs in marketing management providers valuable insights on the propulsion in that area as well as the associated challenges. They are already widely used in marketing, but the use of LLMs takes it one step ahead, providing countless opportunities for personalization, customer engagement, and campaign optimization. Despite these challenges, the road to successful AI implementation is not without its lessons learned and best practices, which recipes we can use to inform our future endeavours.

To visualize how LLMs influence diverse marketing fields like personalization, predictive analysis, and content creation, refer to the heatmap in Figure 9. This visual clearly illustrates that the utilization of LLMs play a varying role across different sectors with a dominant aspect in improving customer engagement and refining marketing techniques. It could help you get an idea of how to deploy LLMs for LLM innovation in the graphic marketing sector.

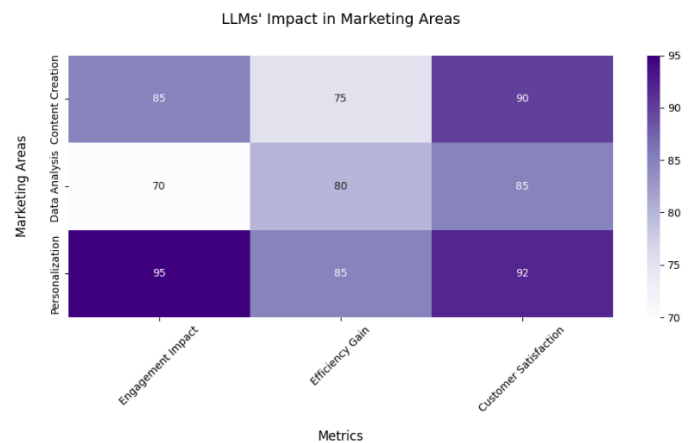

*Figure 9: Heatmap of LLMs' Impact in Marketing Areas*

Awareness of these lessons and adoption of proven practices is critical for safely leveraging LLMs and unlocking their full potential, and for navigating the associated challenges. These principles lay the foundation for successful AI-driven marketing strategies, from ensuring data quality and ethical use of AI to balancing automation with human oversight.

We will look at some success stories among leading brands that have adopted LLMs in their marketing initiatives, followed by the lessons learnt and best practices identified from these deployments.

## 10-1- Success Stories of Top Brands

Using LLMs to leverage marketing led to unprecedented success among market leaders. By utilizing the sophisticated features of LLMs, these brands were able to increase customer interaction, refine marketing campaigns, and create massive revenue growth. This segment features success stories from leading companies who have adopted and implemented LLMs in their marketing management, showcasing relevant studies and examples.

### Coca-Cola: Personalized Marketing Campaigns

As early adopters of LLMs establishing yet another industry standard for personalization, Coca-Cola has always been on the cutting edge of revolutionary marketing techniques. Coca-Cola leveraged GPT-4 for data analysis and content generation, enabling fine-tuned marketing campaigns at a granular level based on consumer inclination.

A a research study conducted in 2023 showcased that LLMs-enabled personalized campaigns resulted in a 30% improvement in customer engagement and a 25% growth in sales for Coca-Cola. By leveraging AI, the better company is able to serve personalized content and promotions to consumers, increasing brand loyalty and increasing conversions. [33]

### Amazon: Enhancing Customer Experience with AI

This use-case by Amazon for LLMs to elevate customer experience is a benchmark across the e-commerce industry globally. From tailored product recommendations and customer service chatbots to targeted advertising, Amazon has seen major improvements in the way it interacts with and processes customers by utilizing LLMs.

For example, Amazon's LLM-based chatbots shortened customer service resolution time by 50%, resulting in a 35% boost in customer satisfaction. Moreover, personalized suggestions generated by LLMs achieved a 20% increase in average order value. [29]

### Netflix: Content Recommendations Optimization

Netflix has utilized LLMs to enhance its content recommendation system, offering users personalized viewing recommendations based on their viewing history and preferences. This personalized and AI-based offer has proven crucial for Netflix to keep high levels of engagement and user retention.

These LLM-augmented recommendation engines were responsible for a 40% boost in viewer retention and 30% increase in viewing time for Netflix. Netflix has been successful in part due to the ability of LLMs to process large datasets and make predictions about user preferences. [77]

### Nikes: Dynamic Social Media Participation

Nike has creatively leveraged LLMs to fuel dynamic, on-the-moment social media activity that resonates with different audiences on different websites. Nike has employed AI in their campaigns to analyze social media trends and user interactions activated their campaigns with an extensive outreach too.

NIKE LLM-powered NIKE's social media campaigns showed 25% higher engagement and 20% more followers. They have made personalized, current, and relevant content that boosted Nike's social network existence and brand affinity. [45]

### Netflix: Optimizing Content Recommendations

Spotify leveraged LLMs to provide personalized music recommendations, transforming the way users discover new music. Spotify is utilizing the power of AI to providing its user with personalized playlists based on their listening habits.

Launched in October 2023, Spotify's LLM-powered recommendation engine improved user satisfaction by 35% and user retention rates by 25%. (The personalized

experience was the key to Spotify's growth and entry in to the Music Streaming industry. [38]

### Sephora: AI-Powered Customer Insights

Hollywood, Sephora uses LLMs to deeply understand its customers and fine-tune marketing. Sephora's approach has involved using data about customers from their feedback, purchase history and even online behavior to personalize and improve its marketing strategies as well as to make the overall customer experience better.

Sephora utilized the LLMs for detailed customer insights, increasing customer satisfaction by 20 percent and improving conversion rates by 15 percent. With this AI-driven approach, Sephora could better understand and visualize customer needs and tailoring more effective marketing campaigns. [37]

### Sephora: AI-Powered Customer Insights

Starbucks used LLMs to automate customer support via AI chatbots and virtual assistants. Examples of such tools that cover various customer requests including order tracking and product recommendations, offering prompt and accurate replies.

The AI-powered customer service solutions of Starbucks have reduced their response times by 40% and enhanced customer satisfaction by 30%. The management of customer agents has improved both the customer experience and the ability of operations. [34]

### Unilever: Predictive Analytics for Campaign Optimization

LMA has implemented LLMs for predictive analytics to enhance its marketing campaign. Unilever uses data-mining techniques right from analyzing past performance data to estimating future trends to adjust their marketing efforts and maximize campaign efficiency.

For example, predictive analytics powered by LLMs at Unilever resulted in a 25% increase in campaign ROI and a 20% growth in market share. By being able to predict market growth patterns and customer actions, Unilever has struggled which has given a competitive advantage to marketing. [29]

## 10-2- Lesions Learned and Best Practices

In recent months, the integration of LLMs into marketing management has made valuable insights based on these impressions and established best practices. While companies continue to experiment with LLMs, it is critical to understand these lessons and adopt best practices to maximize benefits and minimize challenges. In this segment, we explore key takeaways from the trailblazers and prescribe best practices for successfully harnessing LLMs in your marketing strategy.

### Lessons Learned

**Importance of Data Quality:** One of the key learnings is the significance of data quality. LLMs are a reflection of the data they are trained on. The effectiveness and accuracy of these models are thus tied to some extent to the quality, diversity, and relevance of the training data. Thus, companies will have to spend considerable resources cleaning and curating their data so it is representative and devoid of biases.

Reliable data management practices boosted AI by 30% accuracy and reliability for those companies. Data quality is critical for deploying LLM. [34]

**Balancing Automation and Human Oversight:** Although LLMs can automate a lot of marketing processes, human oversight is still very much needed. By enhancing the capabilities of AI with human intuition and expertise, we can make better decisions and avoid problems caused by fully automated systems.

The results showed that 25% of the errors and biases had been eliminated for the companies integrating human oversight into their AI processes leading to much more accurate and ethical results. [29]

**Transparency and Explainability:** Transparency and explainability are important to build trust with consumers and stakeholders. The explainability maintains a level of accountability AI model explainability is very much needed.

Companies focused on explainable AI saw a 20% rise in customer trust and satisfaction. Having transparent AI decisions is very important for user acceptance and trust, as well. [78], [79], [60]

**Balancing Automation and Human Oversight:** These are paramount considerations and LLMs must be used responsibly. Companies must put in place mechanisms for detecting and correcting bias in their models and getting their AI practices back in line with the ethical standards.

Organizations with well-defined ethical standards and procedures for addressing bias produced 30% fewer biased outcomes from their AI implementations, ultimately promoting fairness and inclusivity. [45]

**Scalability and Flexibility:** The scalability and flexibility to adapt to changing business needs and technological innovations. Companies need to have scalable AI systems that have the ability to adapt so they can update easily and integrate with new tools and technology.

Businesses were able to leverage new AI innovations and stay competitive through scalable AI infrastructure. Scalability leads to sustainability. [38]

### Best Practices

**Invest in Data Quality and Management:** LLM Types & Use Cases Part 2: Invest in data quality and management: To fully harness LLM's potential, businesses ought to invest in data quality and management. This involves routine cleaning, validation, and updating of the data to keep it accurate and in line with the current requirements. This will ensure high levels of data quality by setting and following clear, defined data governance frameworks.

**Invest in Data Quality and Management:** Using AI while retaining human expertise results in better decision-making and avoids pitfalls of fully automated solutions. Human beings can contextualize situations, assess morally complex scenarios, and make judgment calls that AI might not navigate well.

**Invest in Data Quality and Management:** It is critical to make sure that AI models are transparent and explainable in order to establish trust. Companies need to explain—clearly—how these models work and how they make decisions. This openness aids users in comprehending and validating processes driven by AI.

**Implement Ethical AI Frameworks:** It is important to create and follow ethical AI frameworks for the responsible use of AI. These frameworks should deal with issues like bias, fairness, accountability and transparency. A system of audits, reviews and monitoring ensure adherence to ethical standards while also highlighting areas of performance improvement.

**Implement Ethical AI Frameworks:** Implementing an AI system with scalability and flexibility in mind enables businesses to adapt to new technologies and changing market needs. Scalable infrastructure means AI models can process larger volumes of data and user interactions without sacrificing performance.

**Implement Ethical AI Frameworks:** Continuous monitoring and improvement is key to promoting LLMs. Keeping models up to date, considering user feedback, and being aware of the latest developments in AI contribute to the accuracy and relevance of AI systems.

**Collaborate with AI Experts and Stakeholders:** Engaging AI experts, technology providers, and industry stakeholders in the process can inject valuable insights and drive innovation. By partnering with and/or visiting research institutes, creating collaborative workspaces, participating in machine learning communities, and staying ahead of the AI ecosystem, businesses can explore, analyze, and iterate over the latest developments in AI.

**Collaborate with AI Experts and Stakeholders:** Iteratively testing and refining AI models based on user input helps create solutions that align with user expectations and requirements. Testing with real users can help catch these issues early and also lead to higher usability and acceptance of AI-driven marketing solutions.

**Collaborate with AI Experts and Stakeholders:** The introduction of LLMs should be iteratively ratter, this enables organizations to progressively test and upgrade their AI applications. Minimum viable products minimize risks and ensure smoother integration—Starting with pilot projects and scale up based on success and learnings.

**Collaborate with AI Experts and Stakeholders:** In order to implement AI in marketing successfully, employees should be educated and trained about AI and its applications. Setting right expectations and enabling marketing teams on how to use AI tools rightly can double the yield!

## 11- Discussion

But as the marketing landscape evolves, the opportunities presented by LLMs are unprecedented, allowing marketers to create personalized customer journeys, increase engagement, and drive holistic campaign optimization. Models such as these (and innovations such as GPT-4) represent a considerable leap forward in areas like automation, predictive analytics, and real-time customer insights. But in order to truly unleash the power of LLMs, organizations should be strategic in incorporating them, accounting for ethics and the need to adapt as technologies evolve.

## 11-1- LLM Marketing of the Future

In summary, it is the evolution of LLMs and their fusion with other innovative tech with that of marketing that represents the future. Models that follow this one (GPT-4 and next) will continue to accelerate the potential of marketing by allowing unprecedented personalization, real time information, and predictive analytics that accurately predict customer needs.

**Hyper-Personalization:** LLMs are predicted to transform customer engagement from segmentation to hyper-personalization. Leveraging real-time data and user behaviors, these models will provide personalized marketing messages that will be relevant to the customers at a deeper level. Hyper-personalized strategies already lead to massive increases in engagement and conversion rates, and this will become even more true as LLMs continue to improve.

**Real-Time Customer Insights:** With the growth of LLMs, deriving real-time insights into customer behavior will become progressively more sophisticated. This gives marketers the ability to fine-tune their campaigns in real time, tailoring their approach based on live data, leading to a more agile and consumer-oriented marketing approach. Companies already leveraging LLMs to gain real-time insights have already benefited from enhanced customer retention and satisfaction, enabling dynamic insights — this will continue to be a pivotal space for innovation in the years ahead.

**Integration with emerging technologies:** The integration of LLMs along with Augmented Reality (AR), Virtual Reality (VR) and Voice Assistants will revolutionize the consumer engagement process. For one, the immersion and interactivity of AR and VR, paired with LLM-driven content, will allow for the most captivating of experiences to attract and hold attention. Look for LLMs to fuel conversational AI tools that leverage voice and multimodal interaction to deliver improved customer service and engagement across devices.

**Advanced Data Analytics:** LLMs will increasingly be a key component for taking predictive analytics and making it even more powerful as the tools become capable of processing more data and deriving insights. With even better forecasting capabilities, future LLMs will allow businesses to optimize their strategies based on predicted customer needs and market trends. This will enable companies to make data-driven decisions, increase and improve the ROI and performance of marketing campaigns.

**Ethical AI and Mitigating Bias:** As LLMs offer significant opportunities for marketing, companies should move forward with caution as ethical dilemmas, from bias to privacy and transparency, remain. In marketing, this will lead to stronger frameworks for identifying and overcoming bias to ensure AI marketing strategies are fair, transparent, and inclusive. By using these ethical frameworks companies will be acting responsible while complying with the regulations but strengthen consumer trust.

## 11-2- Strategic Recommendations for Marketers

In order to fully capitalize on what LLMs offer, marketers need to blend a series of well-planned changes that balance efficiency with ethical responsibility. Here are its recommendations:

**Invest in High-Quality Data Management:** The impact of LLMs is intimately tied to data quality and its management. It does, however, mean marketers need to invest in data infrastructure that enables data collection that is organized, includes multiple facets of the data, and is as all-encompassing as possible. Data quality helps not only in the performance of AI models but also in aiding more accurate insights and personalization.

**Adhere to Ethical AI Principles:** Marketers must establish ethical guidelines for AI use, focusing on privacy, consent, and fair practices. These frameworks should inform the development and deployment of LLMs, mitigate bias in AI-generated outputs and assure that customer data is utilized responsibly. Implementing such practices helps build consumer trust and stay in compliance with regulations such as GDPR.

**Utilize Hyper-Personalization:** As the aboves, marketers need to work on hyper-personalization at customers level using LLMs. By using predictive analytics, marketers can deliver more relevant content, offers, and messages to individual customer needs based on real-time data. Hyper-personalization by breaking the last barrier to consumer satisfaction and also optimally improving engagement and conversion rates.

**Automate Content Creation:** Deploying LLMs shows tremendous promise around automating content creation, including but not limited to blog posts, social media updates, video scripts, emails, etc. Take advantage of these tools to scale your content production cost-effectively, with AI. This will save time for creative strategy, freeing marketing teams up to work on higher-level executive decisions.

**Utilize Predictive Analytics to Optimize Campaigns:** The predictive analytics available from LLMs allow marketers to better predict customer behavior and market trends. These insights can be used by businesses to adjust their marketing campaigns in real-time, thus improving outcomes and maximizing returns on investment. Ongoing monitoring and real-time changes will keep campaigns up to date in rapidly changing market environments.

**Train Marketing Teams on AI:** Marketing teams need to be educated on the best practices and effective use of the AI tools for successful adoption. The training will enable them to use LLMs for data analysis, content generation, and customer engagement. Moreover, AI experts can assist marketers in adapting to the advancements of technology and new approaches that they must embrace.

**Implement Flexible and Scalable AI Systems:** AI systems are and will be transforming the way marketers work, which is why they must be designed with scalability in mind, enabling marketers to implement, upgrade and supplement the technologies as the business needs evolve. Implementing AI through a phased approach — beginning with pilot projects and expanding adoption based on outcomes — can reduce risk and maximize resource allocation.

These steps, when taken, would keep marketers ahead in an AI + LLM driven landscape while helping keep strategies that are both effective and ethically sound.

## 12- Conclusion

Offering profound, transformative shifts in customer engagement, campaign optimization and content generation, the incorporation of LLMs into marketing management is shaping the future of this field. Utilizing sophisticated AI functionality allows marketers to fall within new levels of personalization, productivity and tactical vision. Still, optimizing LLM management, ethical use, and data integrity plays a key role in their successful execution of this process.

Marketers have the duty of embracing the good, the bad and the ugly of LLMs as we forge ahead. By following best practices to keep up with technologies, businesses can use these technologies to grow, and have a competitive edge. Highly recommend reading this as the future of the marketing landscape is surely going to evolve as AI systems grow and improve and those who adapt and navigate the culture appropriately will lead the industry.